\documentclass[10pt,twocolumn,letterpaper]{article}

\usepackage{cvpr}      
\usepackage[dvipsnames]{xcolor}

\definecolor{cvprblue}{rgb}{0.21,0.49,0.74}
\usepackage[pagebackref,breaklinks,colorlinks,citecolor=cvprblue]{hyperref}
\usepackage{graphicx}
\usepackage{subcaption}
\usepackage{float}
\usepackage{booktabs}
\usepackage{hyperref}       
\usepackage{url}            
\usepackage{algorithm} 
\usepackage{algorithmicx}
\usepackage{algpseudocode}

\def\paperID{12580} 
\def\confName{CVPR}
\def\confYear{2024}
\title{Using Human Feedback to Fine-tune Diffusion Models \\without Any Reward Model}

\author{
    Kai Yang$^1$\thanks{Equal contribution. $^\dagger$ Corresponding authors.} \quad
    Jian Tao$^1$\footnotemark[1] \quad
    Jiafei Lyu$^{1\dagger}$ \quad
    Chunjiang Ge$^2$ \quad
    Qimai Li$^3$\\
    Jiaxin Chen$^3$ \quad
    Weihan Shen$^3$ \quad
    Xiaolong Zhu$^3$ \quad
    Xiu Li$^{1\dagger}$ \\
    \textsuperscript{1} Tsinghua Shenzhen International Graduate School, Tsinghua University \\
    \textsuperscript{2} Department of Automation, Tsinghua University \quad
    \textsuperscript{3} Parametrix Technology Company Ltd. \\
    {\tt\small \{yk22,tj22,lvjf20\}@mails.tsinghua.edu.cn} \quad
    {\tt\small li.xiu@sz.tsinghua.edu.cn}
}

\newtheorem{theorem}{Theorem}
\newtheorem{proposition}[theorem]{Proposition}
\begin{document}
\maketitle
\begin{abstract}
Using reinforcement learning with human feedback (RLHF) has shown significant promise in fine-tuning diffusion models. Previous methods start by training a reward model that aligns with human preferences, then leverage RL techniques to fine-tune the underlying models. However, crafting an efficient reward model demands extensive datasets, optimal architecture, and manual hyperparameter tuning, making the process both time and cost-intensive. The direct preference optimization (DPO) method, effective in fine-tuning large language models, eliminates the necessity for a reward model. However, the extensive GPU memory requirement of the diffusion model's denoising process hinders the direct application of the DPO method. To address this issue, we introduce the Direct Preference for Denoising Diffusion Policy Optimization (D3PO) method to directly fine-tune diffusion models. The theoretical analysis demonstrates that although D3PO omits training a reward model, it effectively functions as the optimal reward model trained using human feedback data to guide the learning process. This approach requires no training of a reward model, proving to be more direct, cost-effective, and minimizing computational overhead. In experiments, our method uses the relative scale of objectives as a proxy for human preference, delivering comparable results to methods using ground-truth rewards. Moreover, D3PO demonstrates the ability to reduce image distortion rates and generate safer images, overcoming challenges lacking robust reward models. Our code is publicly available at \href{https://github.com/yk7333/D3PO}{https://github.com/yk7333/D3PO}.
\end{abstract}    
\section{Introduction}

Recent advances in image generation models have yielded unprecedented success in producing high-quality images from textual prompts \cite{ramesh2021zero,saharia2022photorealistic,karras2020analyzing}. Diverse approaches, including Generative Adversarial Networks (GANs) \cite{goodfellow2014generative}, autoregressive models \cite{van2016pixel,ramesh2021zero,ding2021cogview,ding2022cogview2,gafni2022make,esser2021taming}, Normalizing Flows \cite{rezende2015variational,dinh2016density}, and diffusion-based techniques \cite{nichol2021glide,Rombach_2022_CVPR,ramesh2022hierarchical,saharia2022photorealistic}, have rapidly pushed forward the capabilities of these systems. With the proper textual inputs, such models are now adept at crafting images that are not only visually compelling but also semantically coherent, garnering widespread interest for their potential applications and implications.

To adapt the aforementioned models for specific downstream tasks, such as the generation of more visually appealing and aesthetic images, Reinforcement Learning from Human Feedback (RLHF) is commonly employed \cite{christiano2017deep}. This technique has been successfully used to refine large language models such as GPT \cite{brown2020language,openai2023gpt4}. Now, efforts are being made to extend this method to diffusion models to enhance their performance. One such approach, the DDPO method \cite{black2023training}, aims to enhance image complexity, aesthetic quality, and the alignment between prompts and images. The ReLF approach \cite{xu2023imagereward} introduces a novel reward model, named ImageReward, which is specifically trained to discern human aesthetic preferences in text-to-image synthesis. This model is then utilized to fine-tune diffusion models to produce images that align more closely with human preferences. Nonetheless, developing a robust reward model for various tasks can be challenging, often necessitating a vast collection of images and abundant training resources. For example, to diminish the rate of deformities in character images, one must amass a substantial dataset of deformed and non-deformed images generated from identical prompts. Subsequently, a network is constructed to discern and learn the human preference for non-deformed imagery, serving as the reward model.

\begin{figure*}[h]
    \centering
    \includegraphics[width=0.9\linewidth]{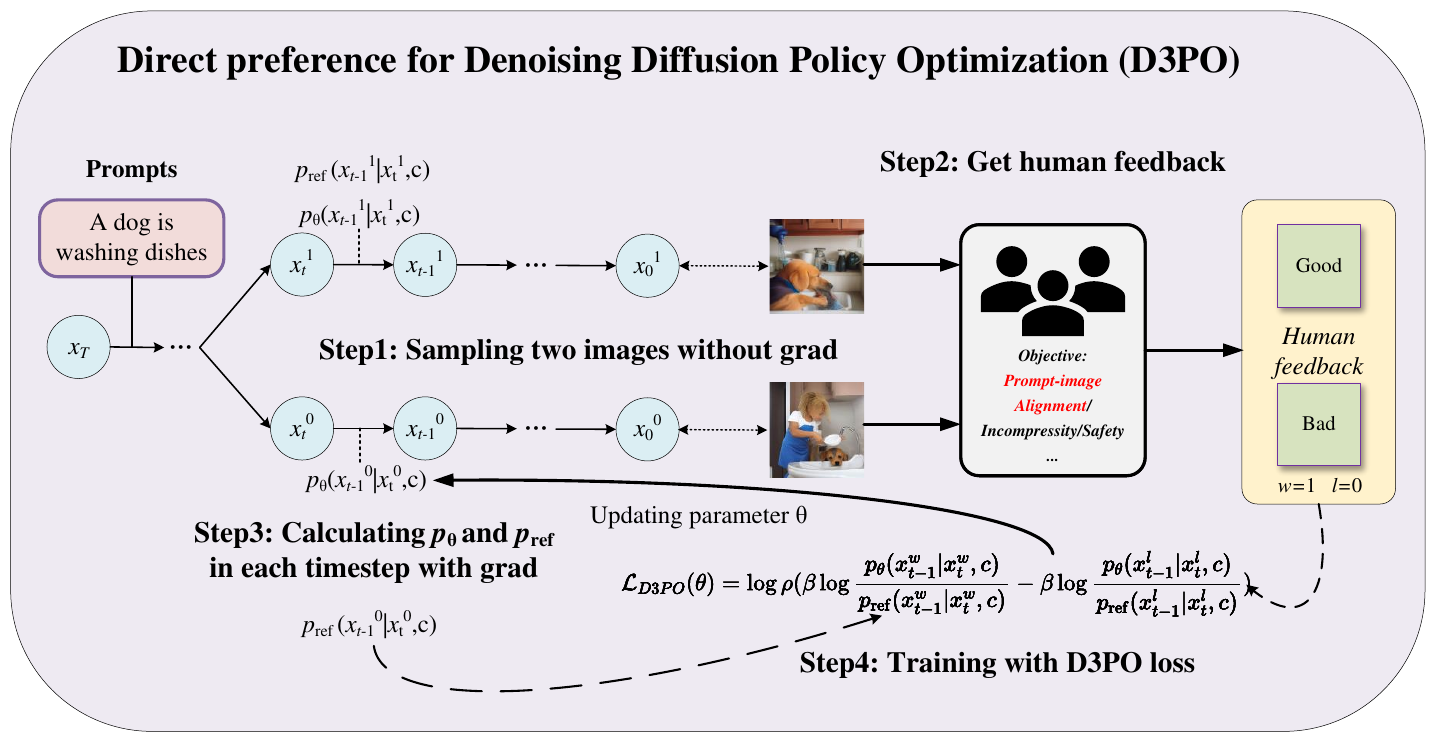}
\caption{Overview of D3PO. The diffusion model generates two corresponding images based on the provided prompts. Guided by specific task requirements—such as improving prompt-image alignment, enhancing image incompressibility, or refining aesthetic quality—human evaluators select the preferred image. Leveraging this human feedback, our method directly updates the diffusion model's parameters without necessitating the training of a reward model.}
    \label{fig:overview}
\end{figure*}

In the field of natural language processing, Direct Preference Optimization (DPO) has been proposed to reduce training costs \cite{rafailov2023direct}. This method forgoes the training of a reward model and directly fine-tunes language models according to human preferences. However, this straightforward and easy-to-train method encounters challenges when applied to fine-tune diffusion models. During the DPO training process, the complete sentence generated by the language model is treated as a single output, necessitating the storage of gradients from multiple forward passes. With diffusion models, one must store the gradients across multiple latent image representations, which are significantly larger than word embeddings, leading to memory consumption that is typically unsustainable.

To address the issue of high computational overhead and enable the use of the DPO method to fine-tune diffusion models directly with human feedback, we conceptualize the denoising process as a multi-step MDP, which utilizes a pre-trained model to represent an action value function $Q$ that is commonly estimated in RL. We extend the theoretical framework of DPO into the formulated MDP, which allows for direct parameter updates at each step of the denoising process based on human feedback, thereby circumventing the significant computational costs and eliminating the need for a reward model. To the best of our knowledge, this is the first work that fine-tune diffusion models without reward models.
Our main contributions are as follows:
\begin{itemize}
    \item We introduce an innovative approach for fine-tuning diffusion models that could significantly modify the current RLHF framework for fine-tuning diffusion models. This method bypasses resource-intensive reward model training by utilizing direct human feedback, making the process more efficient and cost-effective.
    \item We expand the theoretical framework of DPO into a multi-step MDP, demonstrating that directly updating the policy based on human preferences within an MDP is equivalent to learning the optimal reward model first and then using it to guide policy updates. This establishes a robust theoretical foundation and provides assurance for our proposed method.
    \item In our experiments, we have demonstrated the effectiveness of our method by using human feedback to successfully address issues of hand and full-body deformities, enhance the safety of generated images, and improve prompt-image alignment.
\end{itemize}


\section{Related Work}

\textbf{Diffusion models.} 
Denoising diffusion probabilistic models, introduced in \cite{sohl2015deep} and further advanced by \cite{ho2020denoising}, have emerged as powerful tools for generating diverse data types. They have been successfully applied in various domains such as image synthesis \cite{ramesh2021zero, saharia2022photorealistic}, video generation \cite{ho2022imagen, singer2022make, li2023finedance}, and robotics control systems \cite{janner2022planning, ajay2022conditional, chi2023diffusion}. Notably, test-to-image diffusion models have enabled the creation of highly realistic images from textual descriptions \cite{ramesh2021zero, saharia2022photorealistic}, opening new avenues in digital art and design.

Recent studies have focused on refining the guidance of diffusion models for more precise manipulation over the generative process. Techniques such as adapters \cite{zhang2023adding} and compositional approaches \cite{liu2022compositional, du2023reduce} have been introduced to incorporate additional input constraints and blend multiple models, respectively, enhancing image quality and generation control. The implementation of classifier-based \cite{dhariwal2021diffusion} and classifier-free guidance \cite{ho2021classifier} has also contributed significantly to achieving greater autonomy in the generation process, resulting in outputs that closely align with user intentions. In our work, we utilize Stable Diffusion \cite{ramesh2022hierarchical} to generate images based on specific prompts.

\textbf{RLHF.} RLHF stands as a salient strategy in the domain of machine learning when objectives are complex or difficult to define explicitly. This technique has been instrumental across various applications, from gaming, as demonstrated with Atari \cite{christiano2017deep, bai2022training}, to more intricate tasks in robotics \cite{ziegler2019fine, casper2023open}. The integration of RLHF into the development of large language models (LLMs) has marked a significant milestone in the field, with notable models like OpenAI's GPT-4 \cite{openai2023gpt4}, Anthropic's Claude \cite{claude}, Google's Bard \cite{Google}, and Meta's Llama 2-Chat \cite{touvron2023llama} leveraging this approach to enhance their performance and relevance. The effectiveness of RLHF in refining the behavior of LLMs to be more aligned with human values, such as helpfulness and harmlessness, has been extensively studied \cite{bai2022training, ziegler2019fine}. The technique has also proven beneficial in more focused tasks, such as summarization, where models are trained to distill extensive information into concise representations \cite{stiennon2020learning}. Some recent research utilizes Reinforcement Learning from AI Feedback (RLAIF) \cite{lee2023rlaif, zhang2023huatuogpt} as an alternative to RLHF for model fine-tuning. RLAIF offers convenience and efficiency by replacing human feedback with AI-generated feedback. However, for tasks like assessing hand generation normality or image aesthetic appeal, reliable judgment models are currently lacking. Hence, this paper still relies on human feedback for evaluation.

\textbf{Fine-tune Diffusion Models with RL.} Before applying diffusion models, data generation has been regarded as a sequential decision-making problem and combined with reinforcement learning \cite{bachman2015data}. More recently, the SFT method \cite{fan2023optimizing} applied reinforcement learning to diffusion models to enhance existing fast DDPM samplers \cite{ho2020denoising}. Reward Weighted method \cite{lee2023aligning} explored using human feedback to align text-to-image models. It uses the reward model for the coefficients of the loss function instead of using reinforcement learning optimization objectives. ReFL \cite{xu2023imagereward} employs the framework of RLHF. It begins by training a model called ImageReward based on human preferences and then fine-tunes the diffusion model using reinforcement learning. DDPO \cite{black2023training} treats the denoising process of diffusion models as a MDP to fine-tune diffusion models with many reward models. DPOK \cite{fan2023dpok} combine the KL divergence into the DDPO loss and use it to better align text-to-image objectives. All these models need a robust reward model, which demands a substantial dataset of images and extensive human evaluations.

\textbf{Direct Preference Optimization.} In the realm of reinforcement learning, exploring policies derived from preferences rather than explicit rewards has gained attention through various methods. The Contextual Dueling Bandit (CDB) framework \cite{YUE20121538, dudik2015contextual} introduces the concept of a \textit{von Neumann winner}, shifting away from the pursuit of an optimal policy directly based on rewards. \textit{Preference-based Reinforcement Learning} (PbRL) \cite{busa2014preference, pmlr-v206-saha23a,liu2023zero} learns from binary preferences inferred from a cryptic \textit{scoring} function instead of explicit rewards. Recently, the DPO approach \cite{rafailov2023direct} was proposed, which fine-tunes LLMs directly using preferences. DPO leverages the correlation between reward functions and optimal policies, effectively addressing the challenge of constrained reward maximization in a single phase of policy training.

\section{Preliminaries}
\textbf{MDP.} We consider the MDP formulation described in \cite{sutton1998introduction}. In this setting, an agent perceives a state $s\in \mathcal{S}$ and executes an action, where $\mathcal{S}, \mathcal{A}$ denote state space and action space, respectively. The transition probability function, denoted by $P(s'|s, a)$, governs the progression from the current state $s$ to the subsequent state $s'$ upon the agent's action $a$. Concurrently, the agent is awarded a scalar reward $r$, determined by the reward function $r:\mathcal{S}\times\mathcal{A}\to\mathbb{R}$. The agent's objective is to ascertain a policy $\pi(a|s)$ that maximizes the cumulative returns of trajectories $\tau = (s_0,a_0,s_1,a_1,...,s_{T-1},a_{T-1})$, which can be represented as: $\mathcal{J}(\pi) = \mathbb{E}_{\tau}[\sum_{t=0}^{T-1} r\left(s_{t}, a_{t}\right)].$

\textbf{Diffusion models.} Diffusion models learn to model a probability distribution $p(x)$ by inverting a Markovian forward process $q(\boldsymbol{x}_t|\boldsymbol{x}_{t-1})$ which adds noise to the data. The denoising process is modeled by a neural network to predict the mean of $\boldsymbol{x}_{t-1}$ or the noise $\epsilon_{t-1}$ of the forward process. In our work, we use network $\boldsymbol \mu_\theta(\boldsymbol{x}_t;t)$ to predict the mean of $\boldsymbol{x}_{t-1}$ instead of predicting the noise. Using the mean squared error (MSE) as a measure, the objective of this network can be written as:
{\small
\begin{equation}
\mathcal{L}=\mathbb{E}_{t \sim[1, T], \boldsymbol{x}_{0} \sim p\left(\boldsymbol{x}_{0}\right), \boldsymbol{x}_t \sim q(\boldsymbol{x}_t|\boldsymbol{x}_0)}[\left\|\tilde{\boldsymbol{\mu}}(\boldsymbol{x}_0,\boldsymbol{x}_t)-\boldsymbol \mu_{\theta}\left(\boldsymbol{x}_{t}, t\right)\right\|^{2}],
\end{equation}
}
where $\tilde{\boldsymbol{\mu}}_\theta(\boldsymbol{x}_t,\boldsymbol{x}_0)$ is the forward process posterior mean.
In the case of conditional generative modeling, the diffusion models learn to model $p(x|\boldsymbol{c}),$ where $\boldsymbol{c}$ is the conditioning information, i.e., image category and image caption. This is done by adding an additional input $\boldsymbol{c}$ to the denoising neural network, as in $\boldsymbol \mu_\theta(\boldsymbol{x}_t,t;\boldsymbol{c})$. To generate a sample from the learned distribution $p_\theta(x|\boldsymbol{c})$, we start by drawing a sample $\boldsymbol{x}_T\sim \mathcal{N}(\mathbf{0},\mathbf{I})$ and then progressively denoise the sample by iterated application of $\boldsymbol{\epsilon}_\theta$ according to a specific sampler \cite{ho2020denoising}. Given the noise-related parameter $\sigma_t$, the reverse process can be written as:
\begin{equation}\label{denoising}
p_{\theta}\left(\boldsymbol{x}_{t-1} \mid \boldsymbol{x}_{t}, \boldsymbol{c}\right)=\mathcal{N}\left(\boldsymbol{x}_{t-1} ; \boldsymbol{\mu}_{\theta}\left(\boldsymbol{x}_{t}, \boldsymbol{c}, t\right), \sigma_{t}^{2} \mathbf{I}\right).
\end{equation}

\textbf{Reward learning for preferences.} The basic framework to model preferences is to learn a reward function $r^*(s, a)$ from human feedback \cite{wilson2012bayesian,lee2021pebble,christiano2017deep}. The segment $\sigma=\{s_k,a_k,s_{k+1},a_{k+1},...,s_m,a_m\}$ is a sequence of observations and actions. By following the Bradley-Terry model \cite{bradley1952rank}, the human preference distribution $p^*$ by using the reward function can be expressed as:
\begin{equation}\label{prefer distribution}
p^*(\sigma_1 \succ \sigma_0) = \frac{{\exp(\sum_{t=k}^T r^*(s^1_t,a^1_t))}}{\sum_{i\in\{0,1\}} {\exp(\sum_{t=k}^T r^*(s^i_t,a^i_t))}},
\end{equation}
where $\sigma_i \succ \sigma_j$ denotes that segment $i$ is preferable to segment $j$. Now we have the preference distribution of human feedback, and we want to use a network $r_\phi$ to approximate $r^*$. Given the human preference $y\in\{(1,0),(0,1)\}$ which is recorded in dataset $\mathcal{D}$ as a triple $(\sigma_0,\sigma_1,y)$, framing the problem as a binary classification, the reward function modeled as a network is updated by minimizing:
\begin{align}\label{prefer loss}
\mathcal{L}(\phi) = -\mathbb{E}_{(\sigma_1,\sigma_0,y) \sim \mathcal{D}}&[y(0)\log p_\phi(\sigma_0 \succ \sigma_1) \notag \\
&+y(1)\log p_\phi(\sigma_1 \succ \sigma_0)].
\end{align}

\section{Method}
In this section, we describe a method to directly fine-tune diffusion models using human feedback, bypassing the conventional requirement for a reward model. Initially, we reinterpret the denoising process inherent in diffusion models as a multi-step MDP. Then we extend the theory of DPO to MDP, which allows us to apply the principles to effectively translate human preferences into policy improvements in diffusion models.

\subsection{Denoising process as a multi-step MDP}
\label{multiMDP}
We conceptualize the denoising process within the diffusion model as a multi-step MDP, which varies slightly from the approach outlined in \cite{black2023training}. To enhance clarity, we have redefined the states, transition probabilities, and policy functions. The correspondence between notations in the diffusion model and the MDP is established as follows:
{
\small
\begin{align*}
&\mathbf{s}_{t} \triangleq\left(\boldsymbol{c}, t, \boldsymbol{x}_{T-t}\right) \quad P\left(\mathbf{s}_{t+1} \mid \mathbf{s}_{t}, \mathbf{a}_{t}\right) \triangleq\left(\delta_{\boldsymbol{c}}, \delta_{t+1}, \delta_{\boldsymbol{x}_{T-1-t}}\right) \\&\mathbf{a}_{t} \triangleq \boldsymbol{x}_{T-1-t} \quad\quad \quad \pi\left(\mathbf{a}_{t} \mid \mathbf{s}_{t}\right) \triangleq  p_{\theta}\left(\boldsymbol{x}_{T-1-t} \mid \boldsymbol{c}, t , \boldsymbol{x}_{T-t}\right)  
 \\&  \rho_{0}\left(\mathbf{s}_{0}\right) \triangleq\left(p(\boldsymbol{c}), \delta_{0}, \mathcal{N}(\mathbf{0}, \mathbf{I})\right) \;
 \\& r(\mathbf{s}_t,\mathbf{a}_t) \triangleq r(\left(\boldsymbol{c}, t, \boldsymbol{x}_{T-t}\right),\boldsymbol{x}_{T-t-1})
\end{align*}
}
where $\delta_x$ represents the Dirac delta distribution, and $T$ denotes the maximize denoising timesteps. Leveraging this mapping, we can employ RL techniques to fine-tune diffusion models by maximizing returns. However, this approach requires a proficient reward model capable of adequately rewarding the noisy images. The task becomes exceptionally challenging, particularly when $t$ is low, and $\boldsymbol{x}_{T-t}$ closely resembles Gaussian noise, even for an experienced expert.

\subsection{Direct Preference Optimization for MDP}
\label{DPO MDP}
The DPO method does not train a separate reward model but instead directly optimizes the LLMs with the preference data. Given a prompt $x$ and a pair of answers $(y_1,y_0)\sim \pi_\mathrm{ref}(y|x)$, where $\pi_\mathrm{ref}$ represents the reference (pre-trained) model, these responses are ranked and stored in \(\mathcal{D}\) as a tuple \((x, y_w, y_l)\), where \(y_w\) denotes the preferred answer and \(y_l\) indicates the inferior one. DPO optimizes $\pi_\theta$ with the human preference dataset by using the following loss:

\begin{align}\label{DPO loss}
\scalebox{0.82}{$
\mathcal{L}_{\mathrm{DPO}}(\theta) =-\mathbb{E}_{\left(x, y_{w}, y_{l}\right) \sim \mathcal{D}}\left[\log \rho\left(\beta \log \frac{\pi_{\theta}\left(y_{w} \mid x\right)}{\pi_\mathrm {ref }\left(y_{w} \mid x\right)}-\beta \log \frac{\pi_{\theta}\left(y_{l} \mid x\right)}{\pi_\mathrm{ref }\left(y_{l} \mid x\right)}\right)\right].
$}
\end{align}
Here $\rho$ is the logistic function, and $\beta$ is the parameter controlling the deviation from the $\pi_\theta$ and $\pi_\mathrm{ref}$. In our framework, we treat segments $\sigma^1,\sigma^0$ as $y_1,y_0$ and use DPO to fine-tune diffusion models. However, directly using this method faces difficulties since the segments contain a large number (usually 20–50) of the image latent, which occupy a large amount of GPU memory (each image is about 6G even when using LoRA \cite{hu2021lora}). Since we can only get human preferences for the final image $x_0$, if we want to update $\pi_\theta(\sigma)=\prod_{t=k}^T \pi_\theta(s_t,a_t)$, it will consume more than 100G GPU memory, which makes the fine-tuning process nearly impossible.

To address this problem, we extend the DPO theory to MDP. Firstly, we need to reconsider the objective of the RL method. For the MDP problem, the agents take action by considering maximizing the expected return instead of the current reward. For actor-critic methods such as DDPG \cite{lillicrap2015continuous}, the optimization objective of policy $\pi$ gives:
\begin{equation}
\max_\pi\mathbb{E}_{s\sim d^\pi,a\sim \pi(\cdot|s)}[Q^*(s,a)].
\end{equation}
Here, $d^\pi=(1-\gamma) \sum_{t=0}^{\infty} \gamma^{t} P_{t}^{\pi}(s)$ represents the state visitation distribution, where $P_{t}^{\pi}(s)$ denotes the probability of being in state $s$ at timestep $t$ given policy $\pi$. Additionally, $Q^*(s,a)$ denotes the optimal action-value function. The optimal policy can be written as:
\begin{equation}
\pi^*(a|s)=\begin{cases}
1, & \text{if } a = \arg\max_{\hat{a}} Q^*(s,\hat{a}), \\
0, & \text{otherwise}.
\end{cases}
\end{equation}

\begin{figure*}[t]
    \centering
    \includegraphics[width=0.9\textwidth]{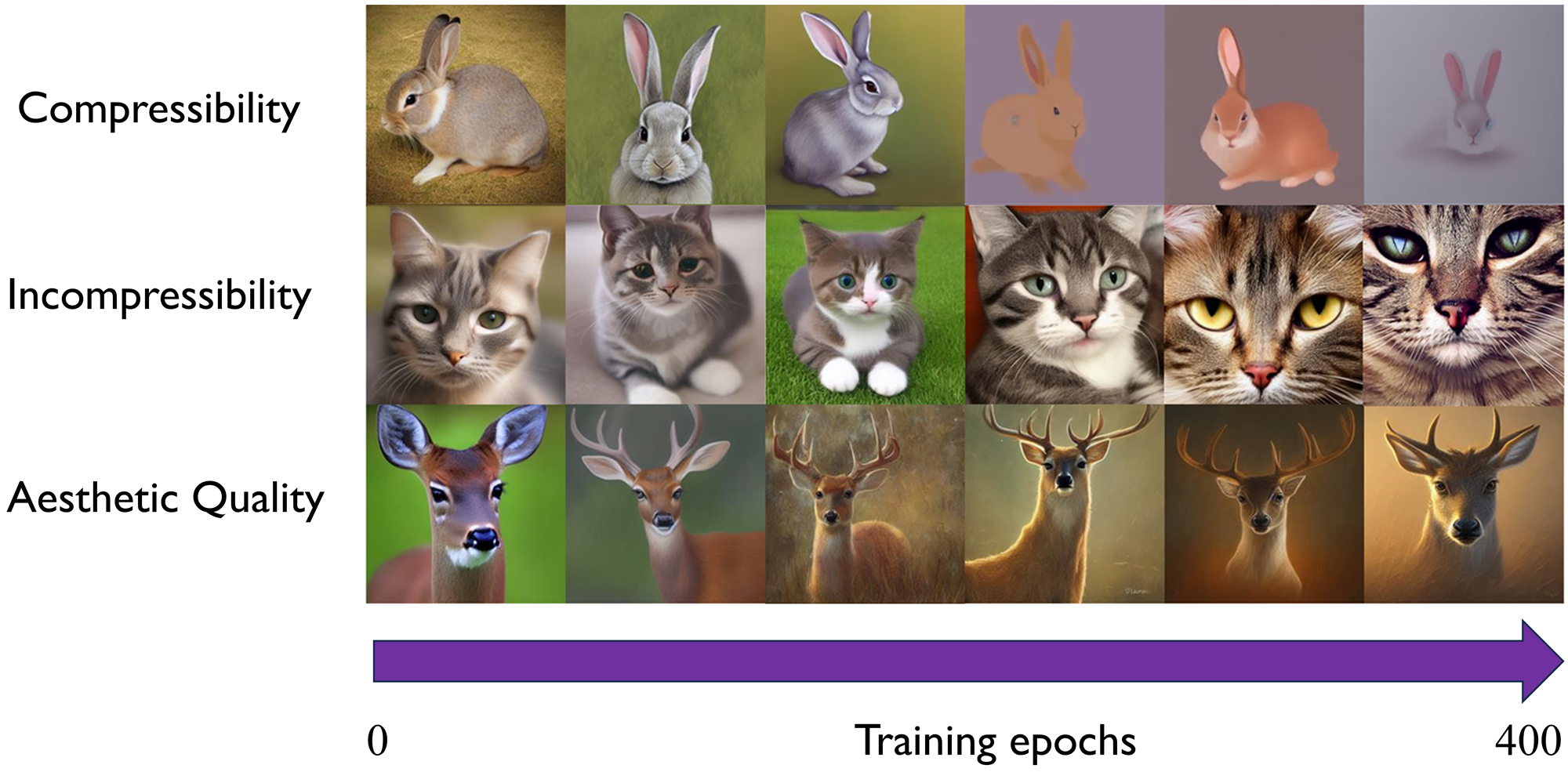}
\caption{Progression of samples targeting compressibility, incompressibility, and aesthetic quality objectives. With the respective focus during training, images exhibit reduced detail and simpler backgrounds for compressibility, richer texture details for incompressibility, and an overall increased aesthetic appeal when prioritizing aesthetic quality.}
    \label{progression}
\end{figure*}

Similar to some popular methods, we use KL-divergence to prevent the fine-tuned policy from deviating from the reference policy, hence relieving the out-of-distribution (OOD) issue. By incorporating this constraint, the RL objective can be rewritten as:
{\small
\begin{equation}\label{RL objective}
\max_\pi\mathbb{E}_{s\sim d^\pi,a\sim \pi(\cdot|s)}[Q^*(s,a)]-\beta \mathbb{D}_{KL}[\pi(a|s)\|\pi_\mathrm{ref}(a|s)].
\end{equation}
}
Here, $\beta$ is the temperature parameter that controls the deviation of $\pi_\theta(a|s)$ and $\pi_\mathrm{ref}(a|s)$.
\begin{proposition}\label{proposition1}
Given the objective of \cref{RL objective}, the optimal policy $\pi^*(a|s)$ has the following expression:
\begin{equation}\label{optimal pi}
\pi^*(a|s)= \pi_\mathrm{ref}(a|s)\exp(\frac{1}{\beta}Q^*(s,a)).
\end{equation}
\end{proposition}

The proof can be seen in Appendix \ref{proof proposition1}. By rearranging the formula of \cref{optimal pi}, we can obtain that:
\begin{equation}\label{Q}
Q^*(s,a) = \beta \log \frac{ \pi^*(a|s)}{\pi_\mathrm{ref}(a|s)}.
\end{equation}
Now, considering \cref{prefer distribution} and noticing that $Q^*(s_t,a_t) = \mathbb{E}\left[\sum_{t=k}^T r^*(s_t,a_t)\right]$ under the policy \( \pi^*(a|s) \), we make a substitution. By replacing \( \sum_{t=k}^T r^*(s_t,a_t) \) with \( Q^*(s_t,a_t) \), we define a new distribution that can be rewritten as:
\begin{equation}\label{prefer Q}
\tilde p^*(\sigma_1 \succ \sigma_0) = \frac{{\exp(Q^*(s_k^1,a_k^1))}}{\sum_{i\in\{0,1\}} {\exp(Q^*(s_k^i,a_k^i))}}.
\end{equation}

We suppose $\sum_{t=k}^m r^*(s_t,a_t)$ is sampled from a normal distribution with mean $\mathbb{E}[\sum_{t=k}^m r^*(s_t,a_t)]$ and standard deviation $\sigma^2$. From a statistical perspective, we can establish the relationship between the new distribution $\tilde{p}^*(\sigma_1 \succ \sigma_0)$ and the raw distribution $p^*(\sigma_1 \succ \sigma_0)$.

\begin{proposition}\label{proposition2}
For \(i \in \{0,1\}\), suppose the expected return satisfies a normal distribution, i.e., \(\sum_{t=0}^{T} r^*\left(s_t^i, a_{t}^i\right) \sim \mathcal{N}\left(Q^{*}(s_0^i,a_0^i), \sigma^2\right)\). Given \(Q^{*}(s,a) \in [Q_\mathrm{\min}, Q_\mathrm{\max}]\) where $Q_\mathrm{\min}$ and $Q_\mathrm{\max}$ represent the lower and upper bounds of the values, then 
    {\small
    \[
    \left| p^*\left(\sigma_{1} \succ \sigma_{0} \right) - \tilde{p}^*\left(\sigma_{1} \succ \sigma_{0} \right) \right| < \frac{(\xi^2 + 1)(\exp(\sigma^2) - 1)}{16\xi \delta}
    \]
    }
with probability at least $1-\delta$, where \(\xi = \frac{\exp(Q_\mathrm{\max})}{\exp(Q_\mathrm{\min})}\).
\end{proposition}

The proof can be seen in Appendix \ref{proof proposition2}. In practical applications, as the volume of data increases, it becomes easier to satisfy the assumption of normality. Additionally, we can use clipping operations to constrain the $Q$ values within a certain range, which introduces upper and lower bounds. Therefore, the aforementioned assumption is reasonable. As shown in proposition \ref{proposition2}, their deviation can be bounded at the scale of $\mathcal{O}(\dfrac{\xi}{\delta}(\exp(\sigma^2)-1))$. It is clear that this bound can approach 0 if the $\sigma^2$ is close to 0. In practice, $\sigma^2$ approaches 0 if the $Q$ function can be estimated with a small standard deviation.

By combining \cref{prefer Q}, \cref{prefer loss}, and \cref{Q}, replacing $p^*\left(\sigma_{1} \succ \sigma_{0} \right)$ with $\tilde{p}^*\left(\sigma_{1} \succ \sigma_{0} \right)$, and substituting $\pi^*(s,a)$ with the policy network $\pi_\theta(s,a)$ that requires learning, we derive the following loss function for updating \( \pi_\theta(a|s) \):
\begin{equation}\label{D3PO loss}
\scalebox{0.91}{$
\mathcal{L}(\theta) = -\mathbb{E}_{(s_k,\sigma_w,\sigma_l)}[\log \rho(\beta\log\frac{\pi_\theta(a_k^w|s_k^w) }{\pi_\mathrm{ref}(a_k^w|s_k^w)}-\beta\log\frac{\pi_\theta(a_k^l|s_k^l) }{\pi_\mathrm{ref}(a_k^l|s_k^l)} )]$},
\end{equation}
where \( \sigma_w = \{s_k^w,\:a_k^w,\:s_{k+1}^w,\:a_{k+1}^w,\:...,\:s_T^w,\:a_T^w\} \) denotes the segment preferred over another segment \( \sigma_l = \{s_k^l,\:a_k^l,\:s_{k+1}^l,\:a_{k+1}^l,\:...,\:s_T^l,\:a_T^l\} \).

\begin{figure*}[h]
    \centering
    \begin{minipage}{0.32\linewidth}
        \centering
        \includegraphics[height=110pt,width=160pt]{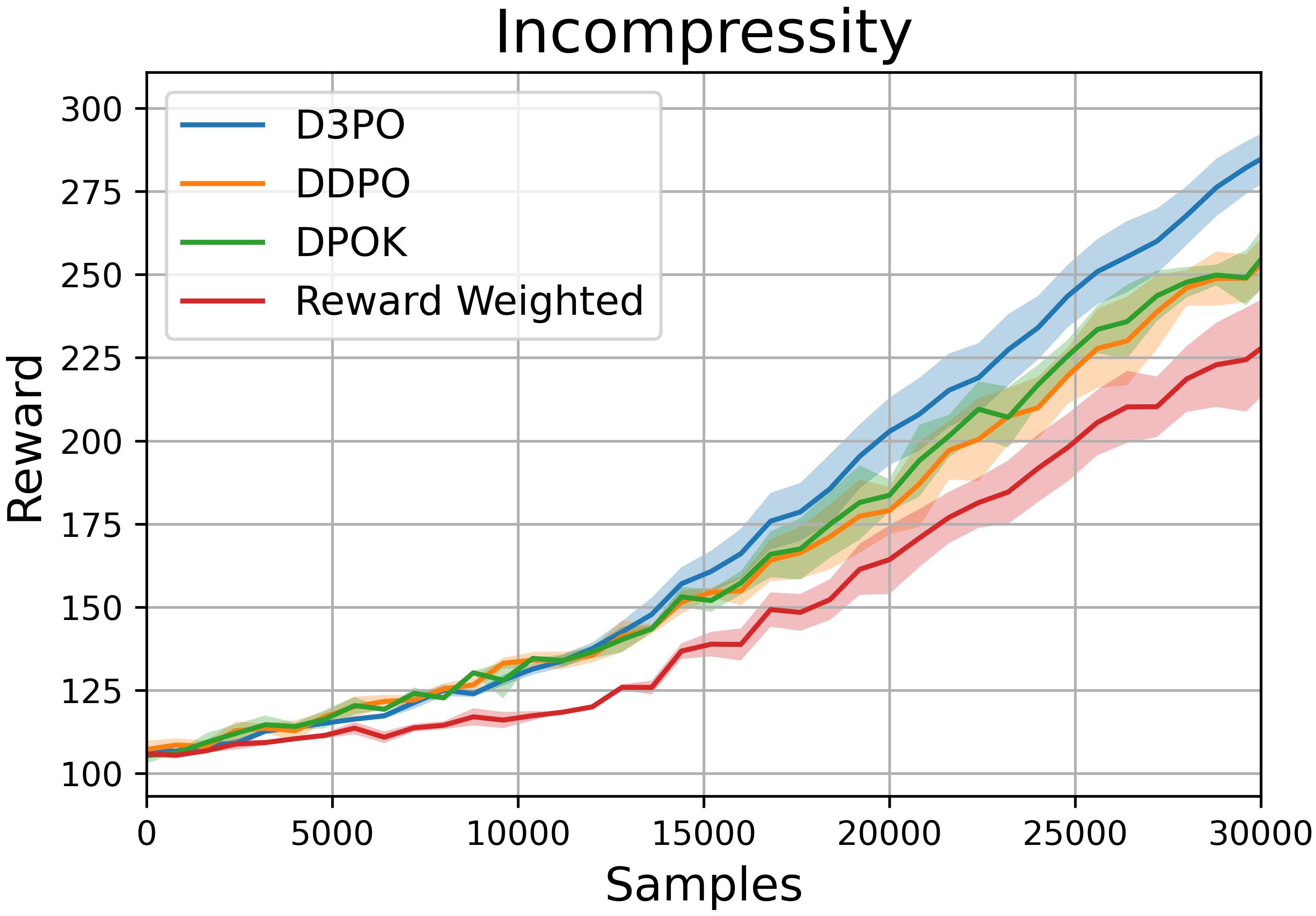}
    \end{minipage}%
    \begin{minipage}{0.32\linewidth}
        \centering
        \includegraphics[height=110pt,width=155pt]{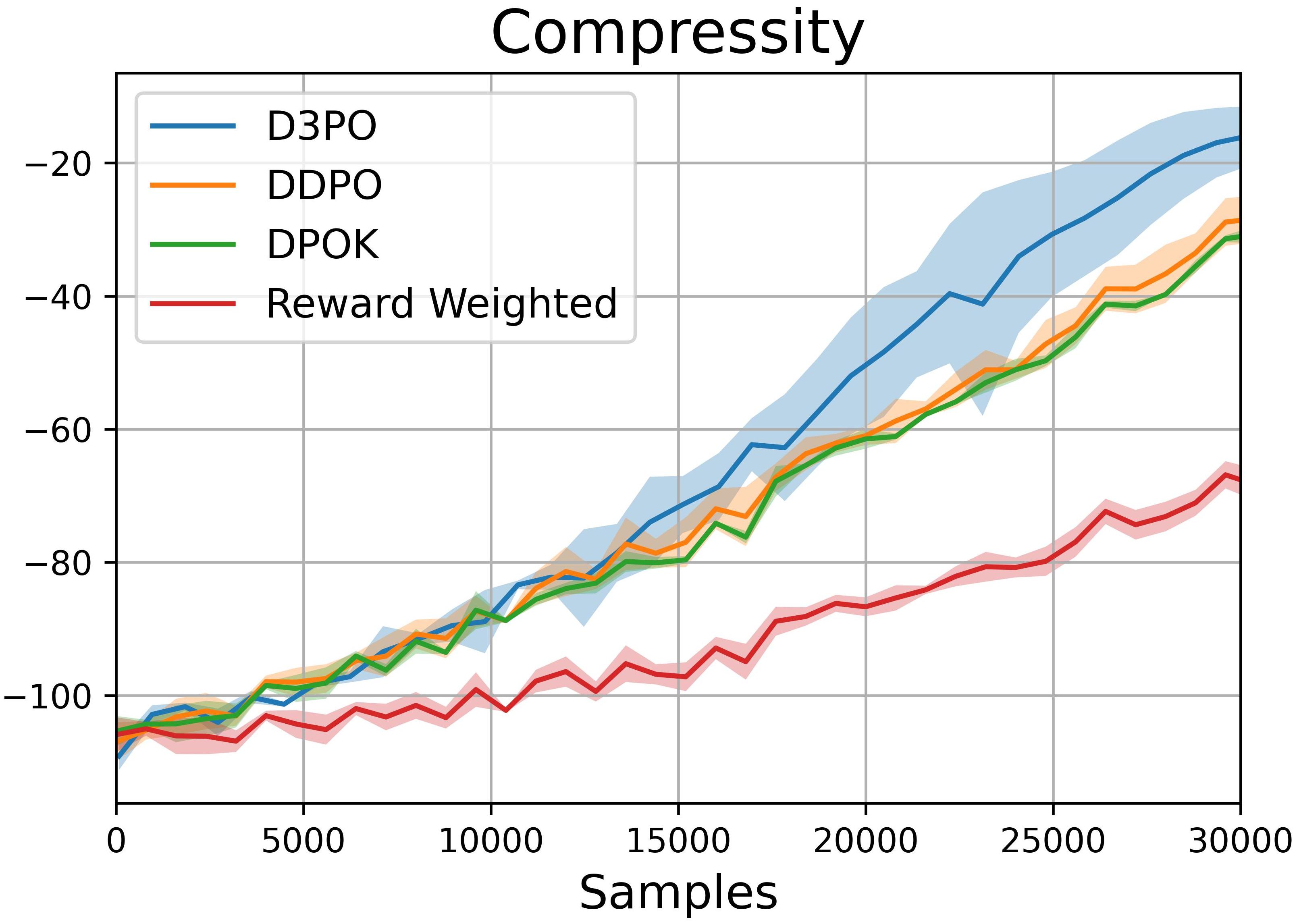}
    \end{minipage}
    \begin{minipage}{0.32\linewidth}
        \centering
        \includegraphics[height=110pt,width=155pt]{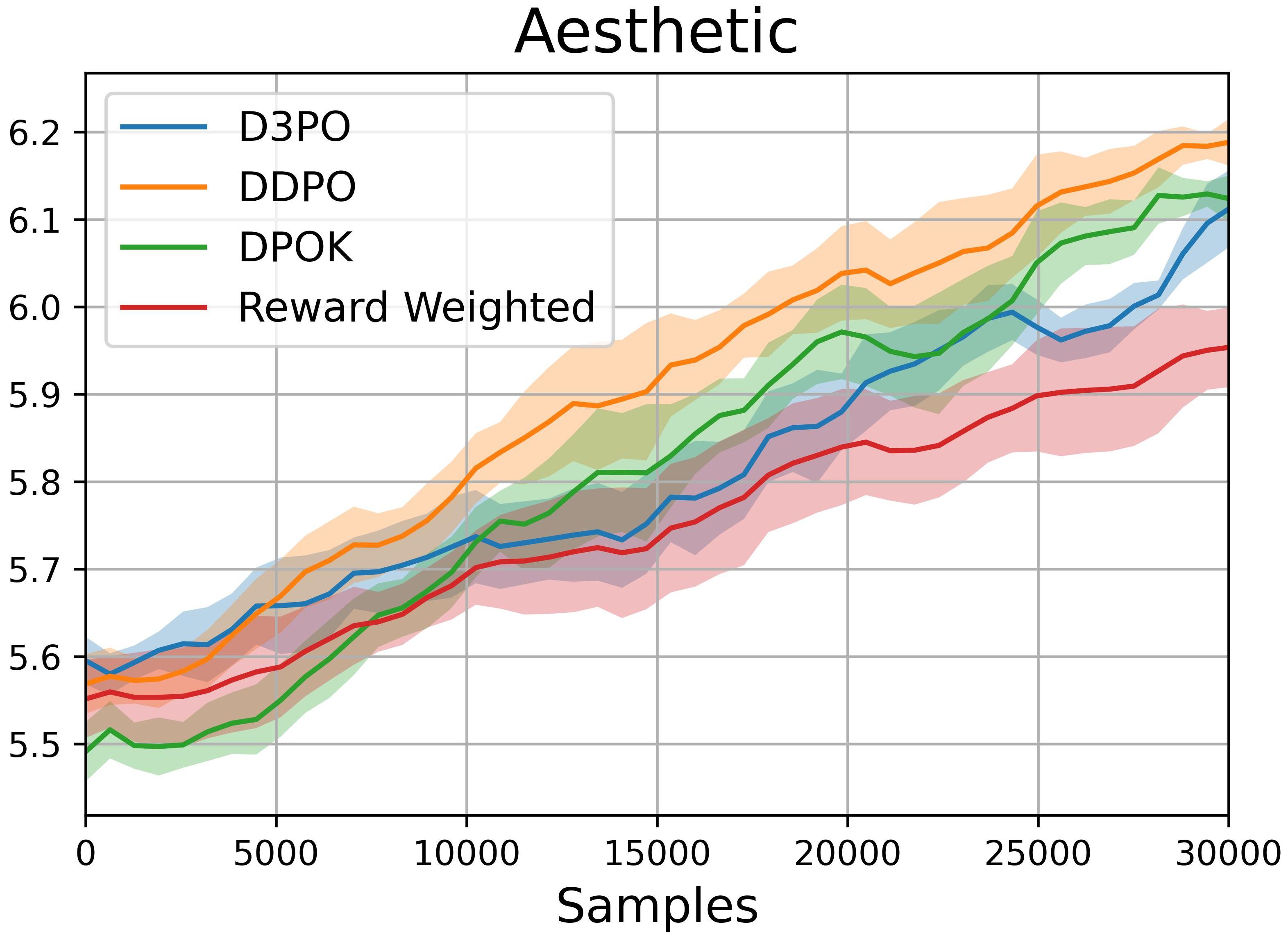}
    \end{minipage}
    \caption{Comparison of D3PO against existing methods. 
The horizontal axis represents the number of image sample pairs generated for updating parameters. The rewards denote image size for incompressity objective, negative image size for the compressity objective, and the LAION aesthetic score for the aesthetic objective. All experiments are conducted with 5 different seeds.}
    \label{curve}
\end{figure*}

\subsection{Direct preference for Denoising Diffusion Policy
Optimization}
Considering the denoising process as a multi-step MDP and using the mapping relationship depicted in Section \ref{multiMDP}, we can use DPO to directly update diffusion models by using \cref{D3PO loss}. In the denoising process, we set $k=0$ and $T=20$. We first sample an initial state $s_0^w=s_0^l=s_0$ and then use \cref{denoising} to generate two segments. After manually choosing which segment is better, the probability of $\pi_\theta(a^w_0|s^w_0)$ is gradually increasing and $\pi_\theta(a^l_0|s^l_0)$ is decreasing, which guides the diffusion model to generate images of human preference. However, the approach of only updating $\pi_\theta(\cdot|s_0)$ does not fully utilize the information within the segment.

Since the middle states of the segment are noises and semi-finished images, it is hard for humans to judge which segment is better by observing the whole segment. But we can conveniently compare the final image $\mathbf{x_0}$. Like many RL methods \cite{silver2016mastering, silver2017mastering, brown2019superhuman} which give rewards by $\forall s_t,a_t\in \sigma, r(s_t,a_t)=1$ for winning the game and $\forall t\in \sigma, r(s_t,a_t)=-1$ for losing the game, we also assume that if the segment is preferred, then any state-action pair of the segment is better than the other segment. By using this assumption, we construct $T$ sub-segments for the agent to learn, which can be written as:
\begin{align}
\scalebox{0.96}{$
    \sigma_i = \{s_i,a_i,s_{i+1},a_{i+1},...,s_{T-1},a_{T-1}\}, \quad 0 \le i \le T-1$}
\end{align}
Using these sub-segments, the overall loss of the D3PO algorithm gives:
\begin{equation}\label{D3PO loss new}
\scalebox{0.84}{$
\mathcal{L}_i(\theta) = -E_{(s_i,\sigma_w,\sigma_l)}[\log \rho(\beta\log\frac{\pi_\theta(a_i^w|s_i^w) }{\pi_\mathrm{ref}(a_i^w|s_i^w)}-\beta\log\frac{\pi_\theta(a_i^l|s_i^l) }{\pi_\mathrm{ref}(a_i^l|s_i^l)} )]$},
\end{equation}
where $i\in[0,T-1]$. Compared to \cref{D3PO loss}, \cref{D3PO loss new} uses every state-action pair for training, effectively increasing the data utilization of the segment by a factor of $T$.

The overview of our method is shown in \cref{fig:overview}. The pseudocode of D3PO can be seen in Appendix \ref{pseudocode}.

\section{Experiment}

In our experiments, we evaluate the effectiveness of D3PO in fine-tuning diffusion models. Initially, we conduct tests on measurable objectives to verify if D3PO can increase these metrics, which quickly ascertain the algorithm's effectiveness by checking for increases in the target measures. Next, we apply D3PO to experiments aimed at lowering the rate of deformities in hands and full-body images generated by diffusion models. Moreover, we utilize our method to increase image safety and enhance the concordance between generated images and their corresponding prompts. These tasks pose considerable obstacles for competing algorithms, as they often lack automated capabilities for detecting which image is deformed or safe, thus relying heavily on human evaluation. We use Stable Diffusion v1.5 \cite{Rombach_2022_CVPR} to generate images in most of the experiments.

\subsection{Pre-defined Quantifiable Objectives Results}
\label{quantifiable objectives}

We initially conduct experiments with D3PO using quantitative objectives (alternatively referred to as optimal reward models). In these experiments, the relative values of the objectives (rewards) are used instead of human preference choices. Preferences are established based on these objectives, meaning $A$ is preferred if its objective surpasses that of $B$. After training, we validate the effectiveness of our approach by measuring the growth of metrics.

In our experimental setup, we benchmark against several popular methods: DDPO \cite{black2023training}, DPOK \cite{fan2023dpok}, and Reward Weighted \cite{lee2023aligning}, all of which necessitate a reward model. Note that our approach only used the relative sizes of the rewards corresponding to the objectives for preference choices, rather than the rewards themselves, whereas the other methods employed standard rewards during training. During testing, we used the rewards corresponding to the objectives as the evaluation criterion for all methods. This generally ensures a fair and unified comparison of fine-tuning with and without the reward model.

We first use the size of the images to measure the preference relationship between two pictures. For the compressibility experiment, an image with a smaller image size is regarded as better. Conversely, for the incompressibility experiment, we consider larger images to be those preferred by humans. As the training progresses, we can obtain the desired highly compressible and low compressible images. We then utilize the LAION aesthetics predictor \cite{schuhmann2022laion} to predict the aesthetic rating of images. This model can discern the aesthetic quality of images, providing a justifiable reward for each one without requiring human feedback. The model can generate more aesthetic images after fine-tuning. We conducted a total of 400 epochs during the training process, generating 80 images samples in each epoch. The progression of the training samples is visually presented in Figure \ref{progression}. More quantitative samples are shown in Figure \ref{samples}. The testing curves of D3PO and other methods are shown in Figure \ref{curve}. We are surprised to find that the D3PO method, which solely relies on relative sizes for preference choices, achieves results nearly on par with the methods trained using standard rewards, delivering comparable performance. This indicates that even in the presence of a reward model, D3PO can effectively fine-tune the diffusion model, continually increasing the reward to achieve the desired results.

\subsection{Experiments without Any Reward Model}
We conduct experiments for some objectives without any reward model. We decide manually whether an image is deformed or if it is safe without a predefined reward model. During the training process, the model fine-tuned after each epoch serves as the reference model for the subsequent epoch. For each prompt, we generate 7 images with the same initial state $x_{T} \sim \mathcal{N}(\mathbf{0},\mathbf{I})$.

\subsubsection{Reduce Image Distortion}
\label{Reduce Image Distortion}

\begin{figure}[h]
    \centering
    \includegraphics[width=0.9\linewidth]{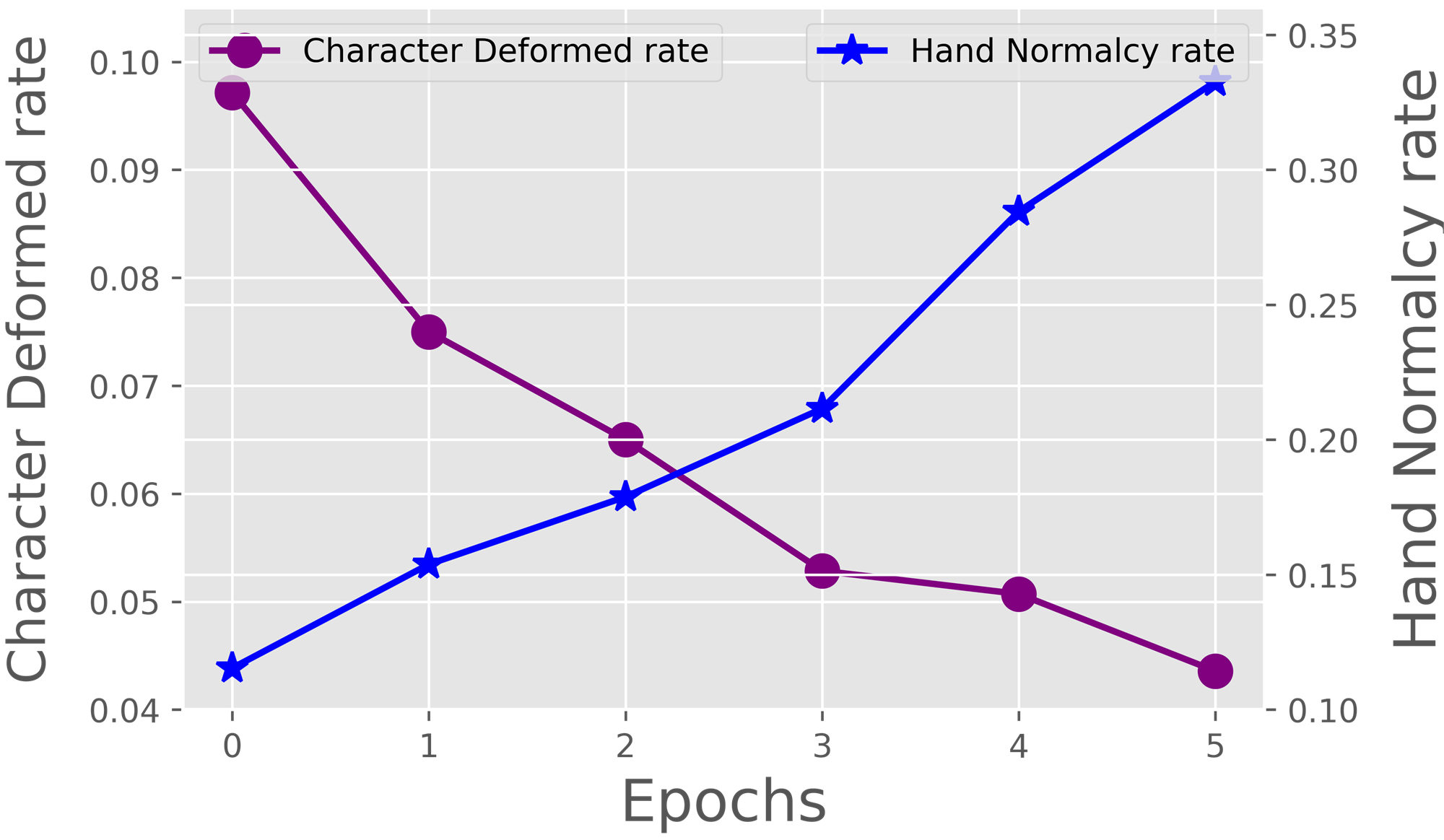}
    \caption{Testing Curves depicting the Normalcy Rate of Hand Images and Deformity Rate of Anime Characters. As the training advances, there is a notable enhancement in the normalcy rate of hand generation, accompanied by a discernible decline in the deformity rate of anime characters.}
    \label{distortion}
\end{figure}

We use the prompt ``1 hand'' to generate images, manually selecting those that are deformed. Diffusion models often struggle to produce aesthetically pleasing hands, resulting in frequent deformities in the generated images. In this experiment, we focus on the normalcy rate of the images instead of the deformity rate, as depicted in Figure \ref{distortion}. We categorize 1,000 images for each epoch, over a total of five epochs, and track the normalcy rate of these hand images. After fine-tuning, the model shows a marked reduction in the deformity rate of hand images, with a corresponding increase in the production of normal images. Additionally, the fine-tuned model shows a higher probability of generating hands with the correct number of fingers than the pre-trained model, as demonstrated in Figure \ref{hand}.

\begin{figure*}[t]
    \centering
    \includegraphics[width=0.9\linewidth]{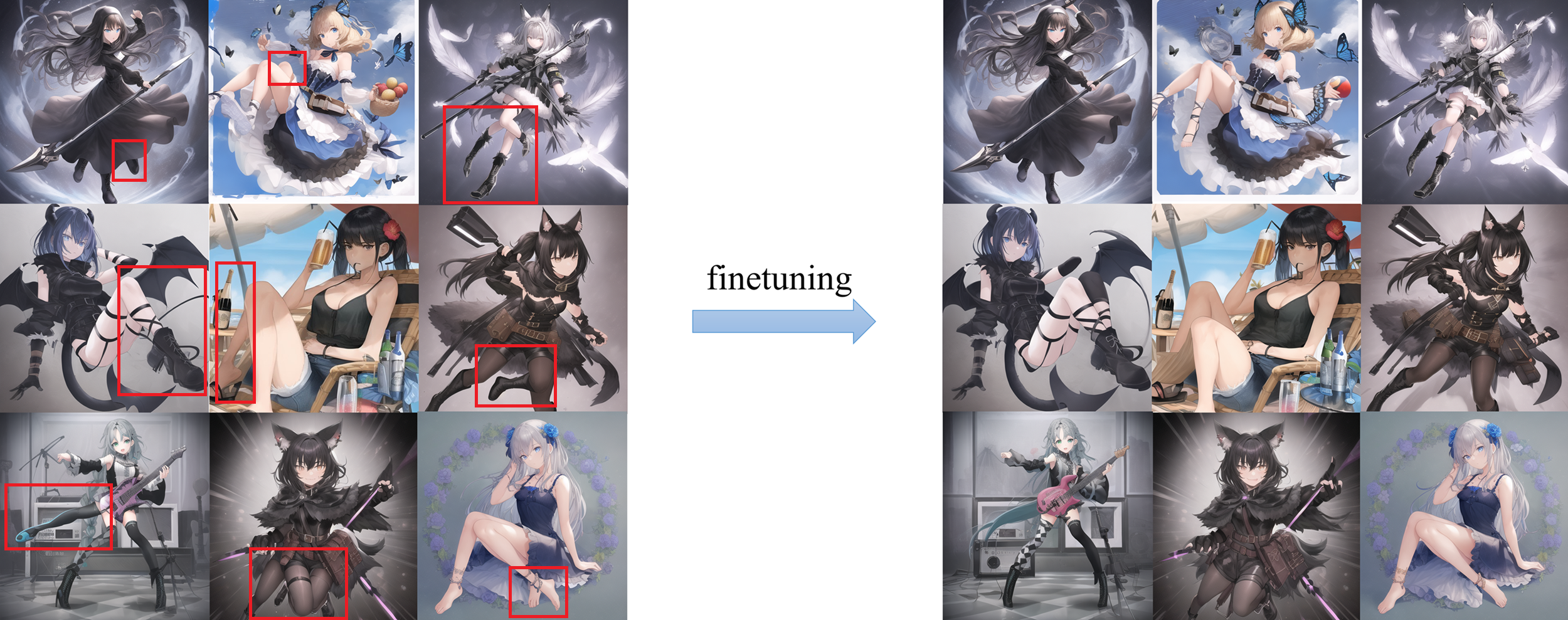}
\caption{Image samples of the pre-trained model and the fine-tuned model. The images on the left side of the arrows are distorted images (such as having 3 legs in the image) generated by the pre-trained model, while the images on the right side are normal images generated after fine-tuning the model. Both sets of images used the same initial Gaussian noise, prompts, and seeds.}
    \label{anything samples}
\end{figure*}

To assess the generality of our method, we generated images with the Anything v5 model \footnote{https://huggingface.co/stablediffusionapi/anything-v5}, renowned for creating anime character images. With Anything v5, there's a risk of generating characters with disproportionate head-to-body ratios or other deformities, such as an incorrect number of limbs (as shown in Figure \ref{anything samples} left). We categorize such outputs as deformed. We assume that non-selected images are more favorable than the deformed ones, though we do not rank preferences within the deformed or non-deformed sets. The diminishing distortion rates across epochs are illustrated in Figure \ref{distortion}, showing a significant decrease initially that stabilizes in later epochs. The visual examples are provided in Figure \ref{anything samples}.

\vspace{-3mm}

\subsubsection{Enhance Image Safety}
\label{Enhance Image Safety}

\begin{figure}[h]
    \centering
    \includegraphics[width=0.75\linewidth]{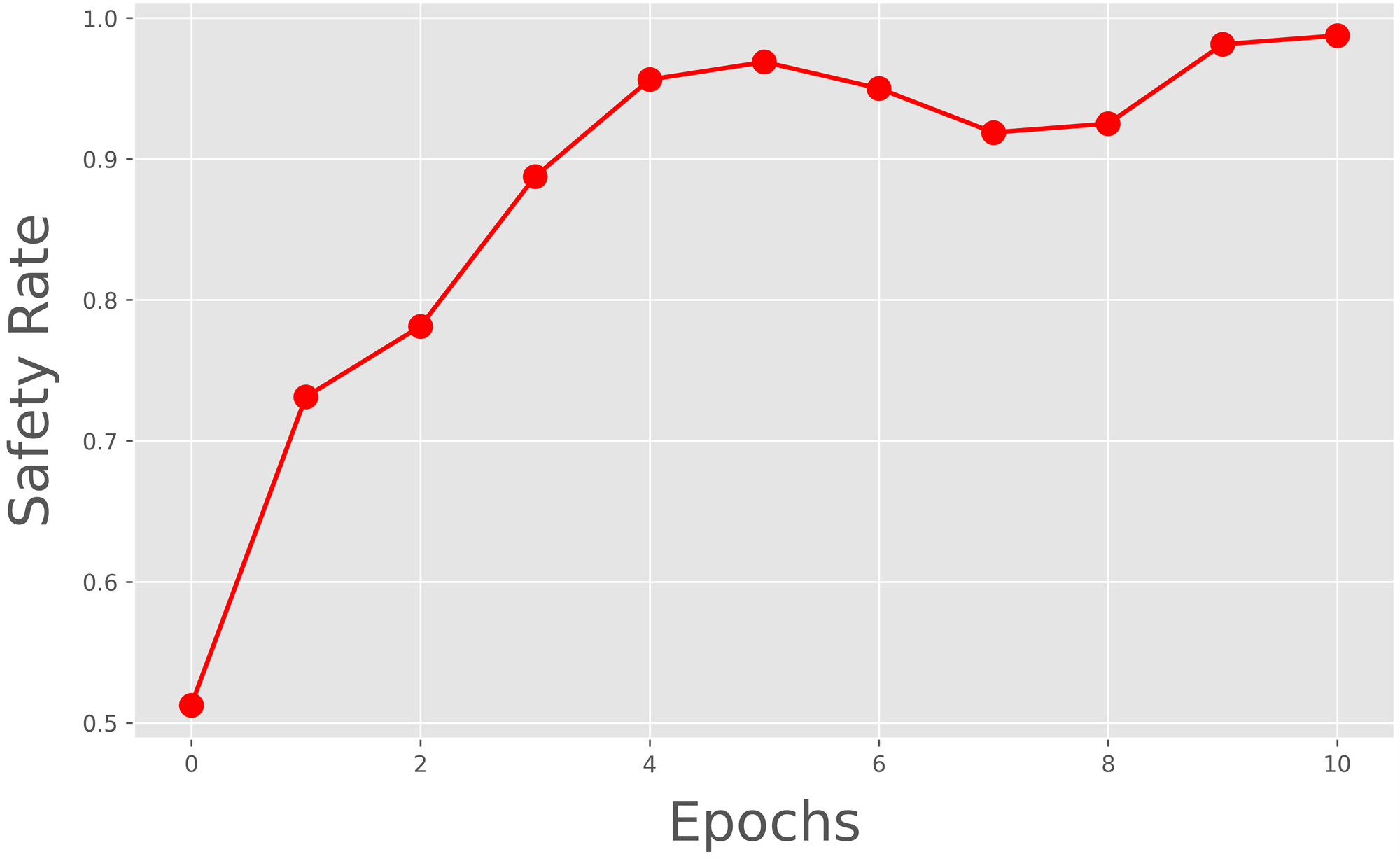}
    \caption{Safety rate curves of the training procession.}
    \label{safety}
\end{figure}

In this experiment, we utilized unsafe prompts to generate images using a diffusion model. These prompts contained edgy terms that could lead to the creation of both normal and Not Safe for Work (NSFW) images, examples being `ambiguous beauty' and `provocative art'. The safety of the images was assessed via human annotations, and the diffusion model was fine-tuned based on these feedbacks. Across 10 epochs, we generated 1,000 images per epoch. Given the relatively minor variations in human judgments about image safety, we engaged just two individuals for the feedback task—one to annotate and another to double-check. The image safety rate during the training process is illustrated in Figure \ref{safety}. After fine-tuning, the model consistently produced safe images, as evidenced in Figure \ref{safe samples}.

\vspace{-4mm}

\begin{figure}[h]
    \centering
    \includegraphics[height=120pt,width=220pt]{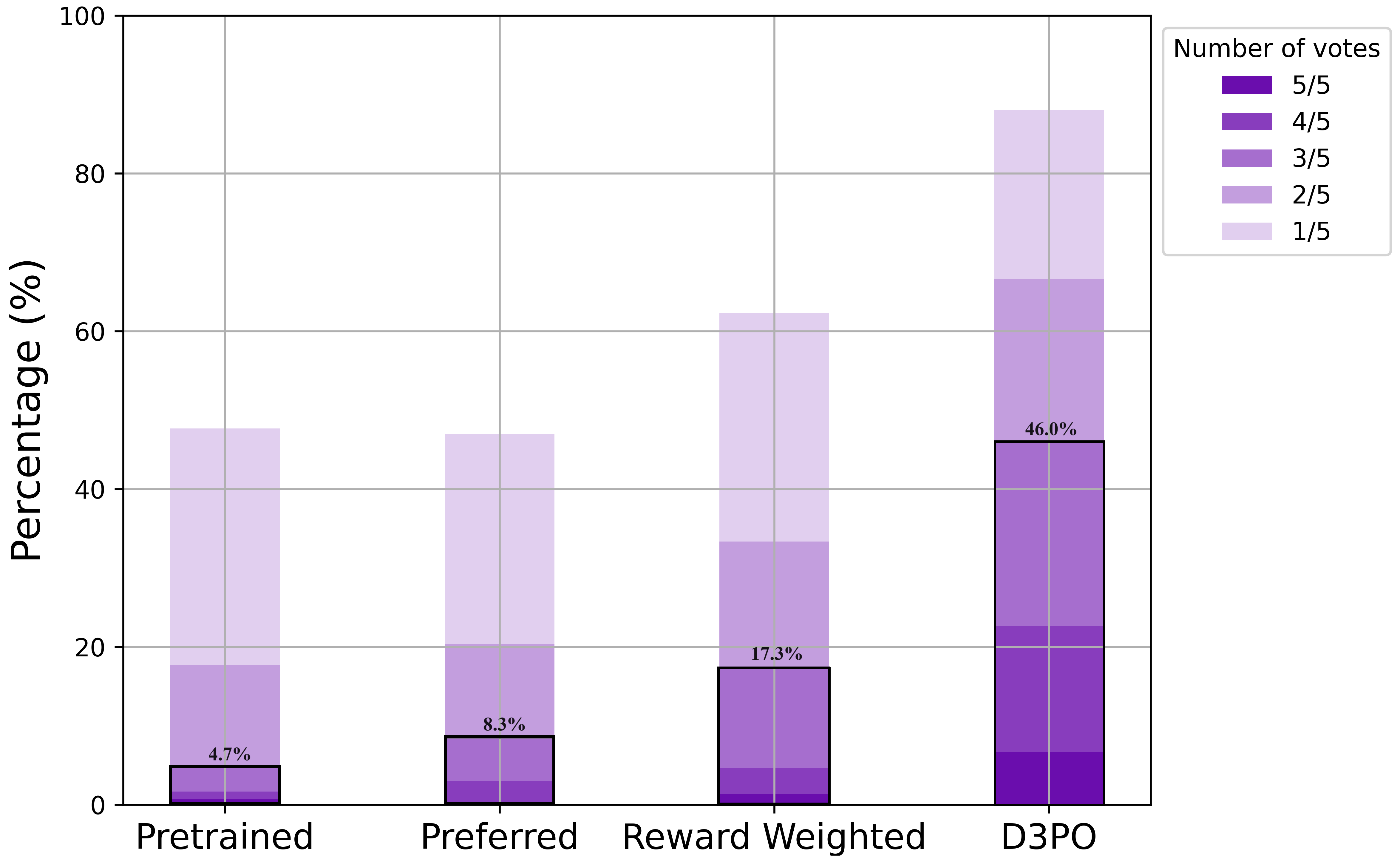}
\caption{Comparative evaluation of 300 text prompts: Our study involved generating images from the pre-trained diffusion model, the fine-tuned models using other methods, and our fine-tuned model—using the same text prompts. For each prompt, human evaluators are tasked with determining which image is better aligned with the text. Each image was assessed by 5 human raters, and we report the percentage of images that received favorable evaluations. We also highlight the percentage with more than half vote in the black box.}
    \label{prefer relationship}
\end{figure}
\vspace{-3mm}

\begin{table}[h]
\centering
\caption{Evaluations of prompt-image alignment. We compare D3PO with those that (a) do not fine-tune, (b) use the preferred images to fine-tune, and (c) use Reward Weighted method to fine-tune. D3PO achieve the best performance. D3PO demonstrated superior performance across all comparisons.}
\label{quantitative table}
\resizebox{\linewidth}{!}{
\begin{tabular}{l|c c c c}
\toprule
\textbf{Methods} & \textbf{CLIP score ($\uparrow$)} & \textbf{BLIP score ($\uparrow$)} & \textbf{ImageReward score ($\uparrow$)} & \textbf{Human preference ($\uparrow$)}\\
\midrule
No fine-tune & 30.7 & 1.95 & 0.04 & 14.7\% \\
Preferred images & 31.0& 1.97& 0.08 & 11\% \\
Reward Weighted & 31.5 & 2.01& 0.17 & 18.7\% \\
D3PO & \textbf{31.9}& \textbf{2.06}& \textbf{0.27}& \textbf{55.7}\% \\
\bottomrule
\end{tabular}
}
\end{table}

\subsubsection{Prompt-Image Alignment} 
\label{sec:prompt}

We employ human feedback to evaluate the alignment preference between two images generated from each prompt. For each epoch, we use 4,000 prompts, generate two images per prompt, and assess preferences with feedback from 16 different evaluators. The training spans 10 epochs. The preference comparisons between images from the pre-trained and fine-tuned models are conducted by an additional 5 evaluators, with the comparative preferences depicted in Figure \ref{prefer relationship}. We also execute quantitative evaluations of models using metrics that measure the congruence between prompts and images, including CLIP \cite{radford2021learning}, BLIP \cite{li2022blip}, and ImageReward \cite{xu2023imagereward}, as presented in Table \ref{quantitative table}.


\section{Conclusion}
In this paper, we propose a direct preference denoising diffusion policy optimization method, named D3PO, to fine-tune diffusion models purely from human feedback without learning a separate reward model. D3PO views the denoising process as a multi-step MDP, making it possible to utilize the DPO-style optimization formula by formulating the action-value function $Q$ with the reference model and the fine-tuned model. D3PO updates parameters at each step of denoising and consumes much fewer GPU memory overheads than directly applying the DPO algorithm. The empirical experiments illustrate that our method achieves competitive or even better performance compared with a diffusion model fine-tuned with a reward model that is trained with a large amount of images and human preferences in terms of image compressibility, image compressibility and aesthetic quality. We further show that D3PO can also benefit challenging tasks such as reducing image distortion rates, enhancing the safety of the generated images, and aligning prompts and images.

\section*{Acknowledgements}
This work was supported by the STI 2030-Major Projects under Grant 2021ZD0201404. 

{
    \small
    \bibliographystyle{ieeenat_fullname}
    \bibliography{main}

\begin{thebibliography}{64}
\providecommand{\natexlab}[1]{#1}
\providecommand{\url}[1]{\texttt{#1}}
\expandafter\ifx\csname urlstyle\endcsname\relax
  \providecommand{\doi}[1]{doi: #1}\else
  \providecommand{\doi}{doi: \begingroup \urlstyle{rm}\Url}\fi

\bibitem[Ajay et~al.(2022)Ajay, Du, Gupta, Tenenbaum, Jaakkola, and Agrawal]{ajay2022conditional}
Anurag Ajay, Yilun Du, Abhi Gupta, Joshua~B Tenenbaum, Tommi~S Jaakkola, and Pulkit Agrawal.
\newblock Is conditional generative modeling all you need for decision making?
\newblock In \emph{The Eleventh International Conference on Learning Representations}, 2022.

\bibitem[Anthropic(2023)]{claude}
Anthropic.
\newblock Introducing claude, 2023.

\bibitem[Bachman and Precup(2015)]{bachman2015data}
Philip Bachman and Doina Precup.
\newblock Data generation as sequential decision making.
\newblock \emph{Advances in Neural Information Processing Systems}, 28, 2015.

\bibitem[Bai et~al.(2022)Bai, Jones, Ndousse, Askell, Chen, DasSarma, Drain, Fort, Ganguli, Henighan, et~al.]{bai2022training}
Yuntao Bai, Andy Jones, Kamal Ndousse, Amanda Askell, Anna Chen, Nova DasSarma, Dawn Drain, Stanislav Fort, Deep Ganguli, Tom Henighan, et~al.
\newblock Training a helpful and harmless assistant with reinforcement learning from human feedback.
\newblock \emph{arXiv preprint arXiv:2204.05862}, 2022.

\bibitem[Black et~al.(2023)Black, Janner, Du, Kostrikov, and Levine]{black2023training}
Kevin Black, Michael Janner, Yilun Du, Ilya Kostrikov, and Sergey Levine.
\newblock Training diffusion models with reinforcement learning.
\newblock \emph{arXiv preprint arXiv:2305.13301}, 2023.

\bibitem[Bradley and Terry(1952)]{bradley1952rank}
Ralph~Allan Bradley and Milton~E Terry.
\newblock Rank analysis of incomplete block designs: I. the method of paired comparisons.
\newblock \emph{Biometrika}, 39\penalty0 (3/4):\penalty0 324--345, 1952.

\bibitem[Brown and Sandholm(2019)]{brown2019superhuman}
Noam Brown and Tuomas Sandholm.
\newblock Superhuman ai for multiplayer poker.
\newblock \emph{Science}, 365\penalty0 (6456):\penalty0 885--890, 2019.

\bibitem[Brown et~al.(2020)Brown, Mann, Ryder, Subbiah, Kaplan, Dhariwal, Neelakantan, Shyam, Sastry, Askell, et~al.]{brown2020language}
Tom Brown, Benjamin Mann, Nick Ryder, Melanie Subbiah, Jared~D Kaplan, Prafulla Dhariwal, Arvind Neelakantan, Pranav Shyam, Girish Sastry, Amanda Askell, et~al.
\newblock Language models are few-shot learners.
\newblock \emph{Advances in neural information processing systems}, 33:\penalty0 1877--1901, 2020.

\bibitem[Busa-Fekete et~al.(2014)Busa-Fekete, Sz{\"o}r{\'e}nyi, Weng, Cheng, and H{\"u}llermeier]{busa2014preference}
R{\'o}bert Busa-Fekete, Bal{\'a}zs Sz{\"o}r{\'e}nyi, Paul Weng, Weiwei Cheng, and Eyke H{\"u}llermeier.
\newblock Preference-based reinforcement learning: evolutionary direct policy search using a preference-based racing algorithm.
\newblock \emph{Machine learning}, 97:\penalty0 327--351, 2014.

\bibitem[Casper et~al.(2023)Casper, Davies, Shi, Gilbert, Scheurer, Rando, Freedman, Korbak, Lindner, Freire, et~al.]{casper2023open}
Stephen Casper, Xander Davies, Claudia Shi, Thomas~Krendl Gilbert, J{\'e}r{\'e}my Scheurer, Javier Rando, Rachel Freedman, Tomasz Korbak, David Lindner, Pedro Freire, et~al.
\newblock Open problems and fundamental limitations of reinforcement learning from human feedback.
\newblock \emph{arXiv preprint arXiv:2307.15217}, 2023.

\bibitem[Chi et~al.(2023)Chi, Feng, Du, Xu, Cousineau, Burchfiel, and Song]{chi2023diffusion}
Cheng Chi, Siyuan Feng, Yilun Du, Zhenjia Xu, Eric Cousineau, Benjamin Burchfiel, and Shuran Song.
\newblock Diffusion policy: Visuomotor policy learning via action diffusion.
\newblock \emph{arXiv preprint arXiv:2303.04137}, 2023.

\bibitem[Christiano et~al.(2017)Christiano, Leike, Brown, Martic, Legg, and Amodei]{christiano2017deep}
Paul~F Christiano, Jan Leike, Tom Brown, Miljan Martic, Shane Legg, and Dario Amodei.
\newblock Deep reinforcement learning from human preferences.
\newblock \emph{Advances in neural information processing systems}, 30, 2017.

\bibitem[Dhariwal and Nichol(2021)]{dhariwal2021diffusion}
Prafulla Dhariwal and Alexander Nichol.
\newblock Diffusion models beat gans on image synthesis.
\newblock \emph{Advances in neural information processing systems}, 34:\penalty0 8780--8794, 2021.

\bibitem[Ding et~al.(2021)Ding, Yang, Hong, Zheng, Zhou, Yin, Lin, Zou, Shao, Yang, et~al.]{ding2021cogview}
Ming Ding, Zhuoyi Yang, Wenyi Hong, Wendi Zheng, Chang Zhou, Da Yin, Junyang Lin, Xu Zou, Zhou Shao, Hongxia Yang, et~al.
\newblock Cogview: Mastering text-to-image generation via transformers.
\newblock \emph{Advances in Neural Information Processing Systems}, 34:\penalty0 19822--19835, 2021.

\bibitem[Ding et~al.(2022)Ding, Zheng, Hong, and Tang]{ding2022cogview2}
Ming Ding, Wendi Zheng, Wenyi Hong, and Jie Tang.
\newblock Cogview2: Faster and better text-to-image generation via hierarchical transformers.
\newblock \emph{Advances in Neural Information Processing Systems}, 35:\penalty0 16890--16902, 2022.

\bibitem[Dinh et~al.(2016)Dinh, Sohl-Dickstein, and Bengio]{dinh2016density}
Laurent Dinh, Jascha Sohl-Dickstein, and Samy Bengio.
\newblock Density estimation using real nvp.
\newblock \emph{arXiv preprint arXiv:1605.08803}, 2016.

\bibitem[Du et~al.(2023)Du, Durkan, Strudel, Tenenbaum, Dieleman, Fergus, Sohl-Dickstein, Doucet, and Grathwohl]{du2023reduce}
Yilun Du, Conor Durkan, Robin Strudel, Joshua~B Tenenbaum, Sander Dieleman, Rob Fergus, Jascha Sohl-Dickstein, Arnaud Doucet, and Will~Sussman Grathwohl.
\newblock Reduce, reuse, recycle: Compositional generation with energy-based diffusion models and mcmc.
\newblock In \emph{International Conference on Machine Learning}, pages 8489--8510. PMLR, 2023.

\bibitem[Dud{\'\i}k et~al.(2015)Dud{\'\i}k, Hofmann, Schapire, Slivkins, and Zoghi]{dudik2015contextual}
Miroslav Dud{\'\i}k, Katja Hofmann, Robert~E Schapire, Aleksandrs Slivkins, and Masrour Zoghi.
\newblock Contextual dueling bandits.
\newblock In \emph{Conference on Learning Theory}, pages 563--587. PMLR, 2015.

\bibitem[Esser et~al.(2021)Esser, Rombach, and Ommer]{esser2021taming}
Patrick Esser, Robin Rombach, and Bjorn Ommer.
\newblock Taming transformers for high-resolution image synthesis.
\newblock In \emph{Proceedings of the IEEE/CVF conference on computer vision and pattern recognition}, pages 12873--12883, 2021.

\bibitem[Fan and Lee(2023)]{fan2023optimizing}
Ying Fan and Kangwook Lee.
\newblock Optimizing ddpm sampling with shortcut fine-tuning.
\newblock \emph{arXiv preprint arXiv:2301.13362}, 2023.

\bibitem[Fan et~al.(2023)Fan, Watkins, Du, Liu, Ryu, Boutilier, Abbeel, Ghavamzadeh, Lee, and Lee]{fan2023dpok}
Ying Fan, Olivia Watkins, Yuqing Du, Hao Liu, Moonkyung Ryu, Craig Boutilier, Pieter Abbeel, Mohammad Ghavamzadeh, Kangwook Lee, and Kimin Lee.
\newblock Dpok: Reinforcement learning for fine-tuning text-to-image diffusion models.
\newblock \emph{arXiv preprint arXiv:2305.16381}, 2023.

\bibitem[Gafni et~al.(2022)Gafni, Polyak, Ashual, Sheynin, Parikh, and Taigman]{gafni2022make}
Oran Gafni, Adam Polyak, Oron Ashual, Shelly Sheynin, Devi Parikh, and Yaniv Taigman.
\newblock Make-a-scene: Scene-based text-to-image generation with human priors.
\newblock In \emph{European Conference on Computer Vision}, pages 89--106. Springer, 2022.

\bibitem[Goodfellow et~al.(2014)Goodfellow, Pouget-Abadie, Mirza, Xu, Warde-Farley, Ozair, Courville, and Bengio]{goodfellow2014generative}
Ian Goodfellow, Jean Pouget-Abadie, Mehdi Mirza, Bing Xu, David Warde-Farley, Sherjil Ozair, Aaron Courville, and Yoshua Bengio.
\newblock Generative adversarial nets.
\newblock \emph{Advances in neural information processing systems}, 27, 2014.

\bibitem[Google(2023)]{Google}
Google.
\newblock Bard, 2023.

\bibitem[Ho and Salimans(2021)]{ho2021classifier}
Jonathan Ho and Tim Salimans.
\newblock Classifier-free diffusion guidance.
\newblock In \emph{NeurIPS 2021 Workshop on Deep Generative Models and Downstream Applications}, 2021.

\bibitem[Ho et~al.(2020)Ho, Jain, and Abbeel]{ho2020denoising}
Jonathan Ho, Ajay Jain, and Pieter Abbeel.
\newblock Denoising diffusion probabilistic models.
\newblock \emph{Advances in neural information processing systems}, 33:\penalty0 6840--6851, 2020.

\bibitem[Ho et~al.(2022)Ho, Chan, Saharia, Whang, Gao, Gritsenko, Kingma, Poole, Norouzi, Fleet, et~al.]{ho2022imagen}
Jonathan Ho, William Chan, Chitwan Saharia, Jay Whang, Ruiqi Gao, Alexey Gritsenko, Diederik~P Kingma, Ben Poole, Mohammad Norouzi, David~J Fleet, et~al.
\newblock Imagen video: High definition video generation with diffusion models.
\newblock \emph{arXiv preprint arXiv:2210.02303}, 2022.

\bibitem[Hu et~al.(2021)Hu, Shen, Wallis, Allen-Zhu, Li, Wang, Wang, and Chen]{hu2021lora}
Edward~J Hu, Yelong Shen, Phillip Wallis, Zeyuan Allen-Zhu, Yuanzhi Li, Shean Wang, Lu Wang, and Weizhu Chen.
\newblock Lora: Low-rank adaptation of large language models.
\newblock \emph{arXiv preprint arXiv:2106.09685}, 2021.

\bibitem[Janner et~al.(2022)Janner, Du, Tenenbaum, and Levine]{janner2022planning}
Michael Janner, Yilun Du, Joshua Tenenbaum, and Sergey Levine.
\newblock Planning with diffusion for flexible behavior synthesis.
\newblock In \emph{International Conference on Machine Learning}, pages 9902--9915. PMLR, 2022.

\bibitem[Karras et~al.(2020)Karras, Laine, Aittala, Hellsten, Lehtinen, and Aila]{karras2020analyzing}
Tero Karras, Samuli Laine, Miika Aittala, Janne Hellsten, Jaakko Lehtinen, and Timo Aila.
\newblock Analyzing and improving the image quality of stylegan.
\newblock In \emph{Proceedings of the IEEE/CVF conference on computer vision and pattern recognition}, pages 8110--8119, 2020.

\bibitem[Kingma and Ba(2014)]{kingma2014adam}
Diederik~P Kingma and Jimmy Ba.
\newblock Adam: A method for stochastic optimization.
\newblock \emph{arXiv preprint arXiv:1412.6980}, 2014.

\bibitem[Lee et~al.(2023{\natexlab{a}})Lee, Phatale, Mansoor, Lu, Mesnard, Bishop, Carbune, and Rastogi]{lee2023rlaif}
Harrison Lee, Samrat Phatale, Hassan Mansoor, Kellie Lu, Thomas Mesnard, Colton Bishop, Victor Carbune, and Abhinav Rastogi.
\newblock Rlaif: Scaling reinforcement learning from human feedback with ai feedback.
\newblock \emph{arXiv preprint arXiv:2309.00267}, 2023{\natexlab{a}}.

\bibitem[Lee et~al.(2021)Lee, Smith, and Abbeel]{lee2021pebble}
Kimin Lee, Laura Smith, and Pieter Abbeel.
\newblock Pebble: Feedback-efficient interactive reinforcement learning via relabeling experience and unsupervised pre-training.
\newblock \emph{arXiv preprint arXiv:2106.05091}, 2021.

\bibitem[Lee et~al.(2023{\natexlab{b}})Lee, Liu, Ryu, Watkins, Du, Boutilier, Abbeel, Ghavamzadeh, and Gu]{lee2023aligning}
Kimin Lee, Hao Liu, Moonkyung Ryu, Olivia Watkins, Yuqing Du, Craig Boutilier, Pieter Abbeel, Mohammad Ghavamzadeh, and Shixiang~Shane Gu.
\newblock Aligning text-to-image models using human feedback.
\newblock \emph{arXiv preprint arXiv:2302.12192}, 2023{\natexlab{b}}.

\bibitem[Li et~al.(2022)Li, Li, Xiong, and Hoi]{li2022blip}
Junnan Li, Dongxu Li, Caiming Xiong, and Steven Hoi.
\newblock Blip: Bootstrapping language-image pre-training for unified vision-language understanding and generation.
\newblock In \emph{International Conference on Machine Learning}, pages 12888--12900. PMLR, 2022.

\bibitem[Li et~al.(2023)Li, Zhao, Zhang, Su, Ren, Zhang, Tang, and Li]{li2023finedance}
Ronghui Li, Junfan Zhao, Yachao Zhang, Mingyang Su, Zeping Ren, Han Zhang, Yansong Tang, and Xiu Li.
\newblock Finedance: A fine-grained choreography dataset for 3d full body dance generation.
\newblock In \emph{Proceedings of the IEEE/CVF International Conference on Computer Vision}, pages 10234--10243, 2023.

\bibitem[Lillicrap et~al.(2015)Lillicrap, Hunt, Pritzel, Heess, Erez, Tassa, Silver, and Wierstra]{lillicrap2015continuous}
Timothy~P Lillicrap, Jonathan~J Hunt, Alexander Pritzel, Nicolas Heess, Tom Erez, Yuval Tassa, David Silver, and Daan Wierstra.
\newblock Continuous control with deep reinforcement learning.
\newblock \emph{arXiv preprint arXiv:1509.02971}, 2015.

\bibitem[Liu et~al.(2022)Liu, Li, Du, Torralba, and Tenenbaum]{liu2022compositional}
Nan Liu, Shuang Li, Yilun Du, Antonio Torralba, and Joshua~B Tenenbaum.
\newblock Compositional visual generation with composable diffusion models.
\newblock In \emph{European Conference on Computer Vision}, pages 423--439. Springer, 2022.

\bibitem[Liu et~al.(2023)Liu, Du, Bai, Lyu, and Li]{liu2023zero}
Runze Liu, Yali Du, Fengshuo Bai, Jiafei Lyu, and Xiu Li.
\newblock Zero-shot preference learning for offline rl via optimal transport.
\newblock \emph{arXiv preprint arXiv:2306.03615}, 2023.

\bibitem[Nichol et~al.(2021)Nichol, Dhariwal, Ramesh, Shyam, Mishkin, McGrew, Sutskever, and Chen]{nichol2021glide}
Alex Nichol, Prafulla Dhariwal, Aditya Ramesh, Pranav Shyam, Pamela Mishkin, Bob McGrew, Ilya Sutskever, and Mark Chen.
\newblock Glide: Towards photorealistic image generation and editing with text-guided diffusion models.
\newblock \emph{arXiv preprint arXiv:2112.10741}, 2021.

\bibitem[OpenAI(2023)]{openai2023gpt4}
OpenAI.
\newblock Gpt-4 technical report, 2023.

\bibitem[Radford et~al.(2021)Radford, Kim, Hallacy, Ramesh, Goh, Agarwal, Sastry, Askell, Mishkin, Clark, et~al.]{radford2021learning}
Alec Radford, Jong~Wook Kim, Chris Hallacy, Aditya Ramesh, Gabriel Goh, Sandhini Agarwal, Girish Sastry, Amanda Askell, Pamela Mishkin, Jack Clark, et~al.
\newblock Learning transferable visual models from natural language supervision.
\newblock In \emph{International conference on machine learning}, pages 8748--8763. PMLR, 2021.

\bibitem[Rafailov et~al.(2023)Rafailov, Sharma, Mitchell, Ermon, Manning, and Finn]{rafailov2023direct}
Rafael Rafailov, Archit Sharma, Eric Mitchell, Stefano Ermon, Christopher~D Manning, and Chelsea Finn.
\newblock Direct preference optimization: Your language model is secretly a reward model.
\newblock \emph{arXiv preprint arXiv:2305.18290}, 2023.

\bibitem[Ramesh et~al.(2021)Ramesh, Pavlov, Goh, Gray, Voss, Radford, Chen, and Sutskever]{ramesh2021zero}
Aditya Ramesh, Mikhail Pavlov, Gabriel Goh, Scott Gray, Chelsea Voss, Alec Radford, Mark Chen, and Ilya Sutskever.
\newblock Zero-shot text-to-image generation.
\newblock In \emph{International Conference on Machine Learning}, pages 8821--8831. PMLR, 2021.

\bibitem[Ramesh et~al.(2022)Ramesh, Dhariwal, Nichol, Chu, and Chen]{ramesh2022hierarchical}
Aditya Ramesh, Prafulla Dhariwal, Alex Nichol, Casey Chu, and Mark Chen.
\newblock Hierarchical text-conditional image generation with clip latents.
\newblock \emph{arXiv preprint arXiv:2204.06125}, 1\penalty0 (2):\penalty0 3, 2022.

\bibitem[Rezende and Mohamed(2015)]{rezende2015variational}
Danilo Rezende and Shakir Mohamed.
\newblock Variational inference with normalizing flows.
\newblock In \emph{International conference on machine learning}, pages 1530--1538. PMLR, 2015.

\bibitem[Rombach et~al.(2022)Rombach, Blattmann, Lorenz, Esser, and Ommer]{Rombach_2022_CVPR}
Robin Rombach, Andreas Blattmann, Dominik Lorenz, Patrick Esser, and Bj\"orn Ommer.
\newblock High-resolution image synthesis with latent diffusion models.
\newblock In \emph{Proceedings of the IEEE/CVF Conference on Computer Vision and Pattern Recognition (CVPR)}, pages 10684--10695, 2022.

\bibitem[Saha et~al.(2023)Saha, Pacchiano, and Lee]{pmlr-v206-saha23a}
Aadirupa Saha, Aldo Pacchiano, and Jonathan Lee.
\newblock Dueling rl: Reinforcement learning with trajectory preferences.
\newblock In \emph{Proceedings of The 26th International Conference on Artificial Intelligence and Statistics}, pages 6263--6289. PMLR, 2023.

\bibitem[Saharia et~al.(2022)Saharia, Chan, Saxena, Li, Whang, Denton, Ghasemipour, Gontijo~Lopes, Karagol~Ayan, Salimans, et~al.]{saharia2022photorealistic}
Chitwan Saharia, William Chan, Saurabh Saxena, Lala Li, Jay Whang, Emily~L Denton, Kamyar Ghasemipour, Raphael Gontijo~Lopes, Burcu Karagol~Ayan, Tim Salimans, et~al.
\newblock Photorealistic text-to-image diffusion models with deep language understanding.
\newblock \emph{Advances in Neural Information Processing Systems}, 35:\penalty0 36479--36494, 2022.

\bibitem[Schuhmann et~al.(2022)Schuhmann, Beaumont, Vencu, Gordon, Wightman, Cherti, Coombes, Katta, Mullis, Wortsman, et~al.]{schuhmann2022laion}
Christoph Schuhmann, Romain Beaumont, Richard Vencu, Cade Gordon, Ross Wightman, Mehdi Cherti, Theo Coombes, Aarush Katta, Clayton Mullis, Mitchell Wortsman, et~al.
\newblock Laion-5b: An open large-scale dataset for training next generation image-text models.
\newblock \emph{Advances in Neural Information Processing Systems}, 35:\penalty0 25278--25294, 2022.

\bibitem[Silver et~al.(2016)Silver, Huang, Maddison, Guez, Sifre, Van Den~Driessche, Schrittwieser, Antonoglou, Panneershelvam, Lanctot, et~al.]{silver2016mastering}
David Silver, Aja Huang, Chris~J Maddison, Arthur Guez, Laurent Sifre, George Van Den~Driessche, Julian Schrittwieser, Ioannis Antonoglou, Veda Panneershelvam, Marc Lanctot, et~al.
\newblock Mastering the game of go with deep neural networks and tree search.
\newblock \emph{nature}, 529\penalty0 (7587):\penalty0 484--489, 2016.

\bibitem[Silver et~al.(2017)Silver, Schrittwieser, Simonyan, Antonoglou, Huang, Guez, Hubert, Baker, Lai, Bolton, et~al.]{silver2017mastering}
David Silver, Julian Schrittwieser, Karen Simonyan, Ioannis Antonoglou, Aja Huang, Arthur Guez, Thomas Hubert, Lucas Baker, Matthew Lai, Adrian Bolton, et~al.
\newblock Mastering the game of go without human knowledge.
\newblock \emph{nature}, 550\penalty0 (7676):\penalty0 354--359, 2017.

\bibitem[Singer et~al.(2022)Singer, Polyak, Hayes, Yin, An, Zhang, Hu, Yang, Ashual, Gafni, et~al.]{singer2022make}
Uriel Singer, Adam Polyak, Thomas Hayes, Xi Yin, Jie An, Songyang Zhang, Qiyuan Hu, Harry Yang, Oron Ashual, Oran Gafni, et~al.
\newblock Make-a-video: Text-to-video generation without text-video data.
\newblock \emph{arXiv preprint arXiv:2209.14792}, 2022.

\bibitem[Sohl-Dickstein et~al.(2015)Sohl-Dickstein, Weiss, Maheswaranathan, and Ganguli]{sohl2015deep}
Jascha Sohl-Dickstein, Eric Weiss, Niru Maheswaranathan, and Surya Ganguli.
\newblock Deep unsupervised learning using nonequilibrium thermodynamics.
\newblock In \emph{International conference on machine learning}, pages 2256--2265. PMLR, 2015.

\bibitem[Stiennon et~al.(2020)Stiennon, Ouyang, Wu, Ziegler, Lowe, Voss, Radford, Amodei, and Christiano]{stiennon2020learning}
Nisan Stiennon, Long Ouyang, Jeffrey Wu, Daniel Ziegler, Ryan Lowe, Chelsea Voss, Alec Radford, Dario Amodei, and Paul~F Christiano.
\newblock Learning to summarize with human feedback.
\newblock \emph{Advances in Neural Information Processing Systems}, 33:\penalty0 3008--3021, 2020.

\bibitem[Sutton et~al.(1998)Sutton, Barto, et~al.]{sutton1998introduction}
Richard~S Sutton, Andrew~G Barto, et~al.
\newblock Introduction to reinforcement learning.
\newblock 1998.

\bibitem[Touvron et~al.(2023)Touvron, Martin, Stone, Albert, Almahairi, Babaei, Bashlykov, Batra, Bhargava, Bhosale, et~al.]{touvron2023llama}
Hugo Touvron, Louis Martin, Kevin Stone, Peter Albert, Amjad Almahairi, Yasmine Babaei, Nikolay Bashlykov, Soumya Batra, Prajjwal Bhargava, Shruti Bhosale, et~al.
\newblock Llama 2: Open foundation and fine-tuned chat models.
\newblock \emph{arXiv preprint arXiv:2307.09288}, 2023.

\bibitem[Van Den~Oord et~al.(2016)Van Den~Oord, Kalchbrenner, and Kavukcuoglu]{van2016pixel}
A{\"a}ron Van Den~Oord, Nal Kalchbrenner, and Koray Kavukcuoglu.
\newblock Pixel recurrent neural networks.
\newblock In \emph{International conference on machine learning}, pages 1747--1756. PMLR, 2016.

\bibitem[Wilson et~al.(2012)Wilson, Fern, and Tadepalli]{wilson2012bayesian}
Aaron Wilson, Alan Fern, and Prasad Tadepalli.
\newblock A bayesian approach for policy learning from trajectory preference queries.
\newblock \emph{Advances in neural information processing systems}, 25, 2012.

\bibitem[Xu et~al.(2023)Xu, Liu, Wu, Tong, Li, Ding, Tang, and Dong]{xu2023imagereward}
Jiazheng Xu, Xiao Liu, Yuchen Wu, Yuxuan Tong, Qinkai Li, Ming Ding, Jie Tang, and Yuxiao Dong.
\newblock Imagereward: Learning and evaluating human preferences for text-to-image generation.
\newblock \emph{arXiv preprint arXiv:2304.05977}, 2023.

\bibitem[Yue et~al.(2012)Yue, Broder, Kleinberg, and Joachims]{YUE20121538}
Yisong Yue, Josef Broder, Robert Kleinberg, and Thorsten Joachims.
\newblock The k-armed dueling bandits problem.
\newblock \emph{Journal of Computer and System Sciences}, 78\penalty0 (5):\penalty0 1538--1556, 2012.
\newblock JCSS Special Issue: Cloud Computing 2011.

\bibitem[Zhang et~al.(2023{\natexlab{a}})Zhang, Chen, Jiang, Yu, Chen, Li, Chen, Wu, Zhang, Xiao, et~al.]{zhang2023huatuogpt}
Hongbo Zhang, Junying Chen, Feng Jiang, Fei Yu, Zhihong Chen, Jianquan Li, Guiming Chen, Xiangbo Wu, Zhiyi Zhang, Qingying Xiao, et~al.
\newblock Huatuogpt, towards taming language model to be a doctor.
\newblock \emph{arXiv preprint arXiv:2305.15075}, 2023{\natexlab{a}}.

\bibitem[Zhang et~al.(2023{\natexlab{b}})Zhang, Rao, and Agrawala]{zhang2023adding}
Lvmin Zhang, Anyi Rao, and Maneesh Agrawala.
\newblock Adding conditional control to text-to-image diffusion models.
\newblock In \emph{Proceedings of the IEEE/CVF International Conference on Computer Vision}, pages 3836--3847, 2023{\natexlab{b}}.

\bibitem[Ziegler et~al.(2019)Ziegler, Stiennon, Wu, Brown, Radford, Amodei, Christiano, and Irving]{ziegler2019fine}
Daniel~M Ziegler, Nisan Stiennon, Jeffrey Wu, Tom~B Brown, Alec Radford, Dario Amodei, Paul Christiano, and Geoffrey Irving.
\newblock Fine-tuning language models from human preferences.
\newblock \emph{arXiv preprint arXiv:1909.08593}, 2019.

\end{thebibliography}
}

\clearpage
\onecolumn
\appendix
\section{D3PO Pseudo-code}
\label{pseudocode}
The pseudocode of the D3PO method can be seen in Algorithm \ref{online}.

\renewcommand{\thealgorithm}{1}
    \begin{algorithm*}[!htbp]
        \caption{D3PO pseudo-code} 
        \label{online}
        \begin{algorithmic}[1]
            \Require Number of inference timesteps $T$, number of training epochs $N$, number of prompts per epoch $K$, pre-trained diffusion model $\boldsymbol{\epsilon}_\theta$.
            \State Copy a pre-trained diffusion model $\boldsymbol{\epsilon}_\mathrm{ref} = \boldsymbol{\epsilon}_\theta$. Set $\boldsymbol{\epsilon}_{\mathrm{ref}}$ with \texttt{requires\_grad} to \texttt{False}.
            
            \For {$n=1:N$}
                \State \texttt{\# Sample images}
                \For{$k=1:K$}
                    \State Random choose a prompt $\boldsymbol{c_k}$ and sample $\boldsymbol{x}_{T}\sim \mathcal{N}(\mathbf{0},\mathbf{I})$
                    \For{$i=0:1$}
                        \For{$t=T:1$}
                                \State \texttt{no grad}: $\boldsymbol{x}^i_{k,t-1}= \mu(\boldsymbol{x}^i_{k,t},t,\boldsymbol{c_k}) +\sigma_{t} \boldsymbol{z}, \;\;\;\; \boldsymbol{z}\sim \mathcal{N}(\mathbf{0},\mathbf{I})$
                    \EndFor
                \EndFor
            \EndFor
                \State \texttt{\# Get Human Feedback}
                \For{$k=1:K$}
                    \State Get human feedback from $\boldsymbol{c_k}$, $\boldsymbol{x}^0_{k,0}$, and $\boldsymbol{x}^1_{k,0}$.
                    \If{$\boldsymbol{x}^0_{0}$ is better than $\boldsymbol{x}^1_{0}$}
                        \State $h_k=[1,-1]$
                    \ElsIf{$\boldsymbol{x}^0_{1}$ is better than $\boldsymbol{x}^0_{0}$}
                        \State $h_k=[-1,1]$
                    \Else
                        \State $h_k=[0,0]$
                    \EndIf
                \EndFor
                \State \texttt{\# Training}
                \For{$k=1:K$}
                \For{$t=T:1$}
                \For{$i=0:1$}
                \State \texttt{with grad}: \State $\mu_\theta(\boldsymbol{x}^i_{k,t},t,\boldsymbol{c_k})=\frac{1}{\sqrt{\alpha_{t}}}\left(\boldsymbol{x}^i_{k,t}-\frac{\beta_{t}}{\sqrt{1-\bar{\alpha}_{t}}} \boldsymbol{\epsilon}_{\theta}\left(\boldsymbol{x}^i_{k,t}, t,\boldsymbol{c_k}\right)\right)$
                \State $\mu_\mathrm{ref}(\boldsymbol{x}^i_{k,t},t,\boldsymbol{c_k}) =\frac{1}{\sqrt{\alpha_{t}}}\left(\boldsymbol{x}^i_{k,t}-\frac{\beta_{t}}{\sqrt{1-\bar{\alpha}_{t}}} \boldsymbol{\epsilon}_\mathrm{ref}\left(\boldsymbol{x}^i_{k,t}, t,\boldsymbol{c_k}\right)\right)$

                \State $\pi_\theta(\boldsymbol{x}^i_{k,t-1}|\boldsymbol{x}^i_{k,t},t,\boldsymbol{c_k})=\frac{1}{\sqrt{2\pi}\sigma_t}\exp(-\frac{(\boldsymbol{x}^i_{k,t-1}-\mu_\theta(\boldsymbol{x}^i_{k,t},t,\boldsymbol{c_k}))^2}{2\sigma_t^2}) $
                \State $\pi_\text{ref}(\boldsymbol{x}^i_{k,t-1}|\boldsymbol{x}^i_{k,t},t,\boldsymbol{c_k})=\frac{1}{\sqrt{2\pi}\sigma_t}\exp(-\frac{(\boldsymbol{x}^i_{k,t-1}-\mu_\mathrm{ref}(\boldsymbol{x}^i_{k,t},t,\boldsymbol{c_k}))^2}{2\sigma_t^2})$
                \EndFor
                \State Update $\theta$ with gradient descent using
                \[
                \nabla_\theta \log \rho(h_k(0) \beta  \log \frac{\pi_\theta(\boldsymbol{x}^0_{k,t-1}|\boldsymbol{x}^0_{k,t},t,\boldsymbol{c})}{\pi_\mathrm{ref}(\boldsymbol{x}_{k,t-1}|\boldsymbol{x}^0_{k,t},t,\boldsymbol{c})} + h_k(1) \beta  \log \frac{\pi_\theta(\boldsymbol{x}^1_{k,t-1}|\boldsymbol{x}^1_{k,t},t,\boldsymbol{c})}{\pi_\mathrm{ref}(\boldsymbol{x}_{k,t-1}|\boldsymbol{x}^1_{k,t},t,\boldsymbol{c})}) 
                \]
                \EndFor
                
            \EndFor
            \EndFor  
        \end{algorithmic}
    
    \end{algorithm*}

\section{Proof}
\label{proof}
\subsection{Proof of Proposition \ref{proposition1}}
\label{proof proposition1}
The RL objective can be written as:
\begin{align*}
\max_{\pi} & \; \mathbb{E}_{s\sim d^\pi,a\sim \pi(a|s)}[Q^*(s,a)] - \beta \mathbb{D}_{KL}[\pi(a|s)\|\pi_\mathrm{ref}(a|s)]  \\
&= \max_\pi\mathbb{E}_{s\sim d^\pi,a\sim \pi(a|s)}[Q^*(s,a)-\beta\log\dfrac{\pi(a|s)}{\pi_\mathrm{ref}(a|s)}]  \\
& = \min_\pi\mathbb{E}_{s\sim d^\pi,a\sim \pi(a|s)}[\log\dfrac{\pi(a|s)}{\pi_\mathrm{ref}(a|s)}-\dfrac{1}{\beta}Q^*(s,a)]  \\
& = \min_\pi\mathbb{E}_{s\sim d^\pi,a\sim \pi(a|s)}[\log\dfrac{\pi(a|s)}{\pi_\mathrm{ref}(a|s)\exp(\dfrac{1}{\beta}Q^*(s,a))}]\\
& = \min_\pi\mathbb{E}_{s\sim d^\pi}[\mathbb{D}_{KL}[\pi(a|s)\|\tilde \pi(a|s)]]
\end{align*}
where $\tilde \pi(a|s) = \pi_\mathrm{ref}(a|s)\exp(\dfrac{1}{\beta}Q^*(s,a))$. Note that the KL-divergence is minimized at 0 iff the two distributions are identical, so the optimal solution is:
$$
\pi(a|s)= \tilde \pi(a|s) = \pi_\mathrm{ref}(a|s)\exp(\dfrac{1}{\beta}Q^*(s,a)).
$$
\subsection{Proof of Proposition \ref{proposition2}}
\label{proof proposition2}
For simplicity, we define $Q_i=Q^*(s_0^i,a^i_0)$ and $X_i=\sum_{t=0}^{T}r^*\left(s_t^i, a_{t}^i\right) \quad i\in\{0,1\}$. Using the \cref{prefer distribution} we can obtain that:
\begin{align*}
\mathbb{E}[p^*\left(\sigma_{1} \succ \sigma_{0} \right)]&=\dfrac{\mathbb{E}[\exp(X_1)]}{\mathbb{E}[\exp(X_1)+\exp(X_0)]}\\&=\dfrac{\exp(Q_1+1/2\sigma)}{\exp(Q_1+1/2\sigma)+\exp(Q_0+1/2\sigma)}\\&=\dfrac{\exp(Q_1)}{\exp(Q_1)+\exp(Q_0)}\\&=\mathbb{E}[\tilde p^*\left(\sigma_{1} \succ \sigma_{0} \right)].
\end{align*}

\begin{align*}
\mathbb{E}[(p^*\left(\sigma_{1} \succ \sigma_{0} \right))^2]&=\dfrac{\mathbb{E}[\exp(2X_1)]}{\mathbb{E}[\exp(2X_1)]+\mathbb{E}[\exp(2X_0)]+\mathbb{E}[2\exp(X_0)\exp(X_1)]}\\&=\dfrac{\exp(2Q_1+2\sigma^2)}{\exp(2Q_1+2\sigma^2)+\exp(2Q_0+2\sigma^2)+\exp(Q_0+Q_1+\sigma^2)}\\&=
\dfrac{\exp(2Q_1+\sigma^2)}{\exp(2Q_1+\sigma^2)+\exp(2Q_0+\sigma^2)+2\exp(Q_0+Q_1)}.
\end{align*}

\begin{align*}
\mathrm{Var}[p^*(\sigma_{1} \succ \sigma_{0})] &
= \mathbb{E}[(p(\sigma_{1} \succ \sigma_{0}))^2] - (\mathbb{E}[p(\sigma_{1} \succ \sigma_{0})])^2 \\&
= \dfrac{2\exp(3Q_1+Q_0)(\exp(\sigma^2)-1)}{[\exp(2Q_1+\sigma^2)+\exp(2Q_0+\sigma^2)+2\exp(Q_0+Q_1)][\exp(Q_1)+\exp(Q_0)]^2} \\&
\le \dfrac{2\exp(3Q_1+Q_0)(\exp(\sigma^2)-1)}{[\exp(Q_1)+\exp(Q_0)]^4}.
\end{align*}

Similarly, we have:

\begin{align*}
\mathrm{Var}[p^*\left(\sigma_{0} \succ \sigma_{1} \right)]\le\dfrac{2\exp(Q_1+3Q_0)(\exp(\sigma^2)-1)}{[\exp(Q_1)+\exp(Q_0)]^4}.
\end{align*}

Note that $\mathrm{Var}[p^*\left(\sigma_{1} \succ \sigma_{0} \right)]=\mathrm{Var}[1-p^*\left(\sigma_{0} \succ \sigma_{1} \right)]=\mathrm{Var}[p^*\left(\sigma_{0} \succ \sigma_{1} \right)]$, considering these two inequalities, we have:

\begin{align*}
\mathrm{Var}[p^*\left(\sigma_{1} \succ \sigma_{0} \right)] &\le \dfrac{[\exp(Q_1+3Q_0)+\exp(Q_0+3Q_1)](\exp(\sigma^2)-1)}{[\exp(Q_1)+\exp(Q_0)]^4}\\&\le\dfrac{[\exp(Q_1+3Q_0)+\exp(Q_0+3Q_1)](\exp(\sigma^2)-1)}{16[\exp(2Q_1)\exp(2Q_0)]}\\&=\dfrac{[\exp(Q_0-Q_1)+\exp(Q_1-Q_0)](\exp(\sigma^2)-1)}{16}\\&\le\dfrac{(\xi+\dfrac{1}{\xi})(\exp(\sigma^2)-1)}{16}.
\end{align*}

By using the Chebyshev inequality, we can obtain:
$$
P(|p^*\left(\sigma_{1} \succ \sigma_{0} \right)-\tilde p^*\left(\sigma_{1} \succ \sigma_{0} \right)|< t)  >1-\dfrac{(\xi^2+1)(\exp(\sigma^2)-1)}{16\xi t}.
$$

We choose $t=\dfrac{(\xi^2+1)(\exp(\sigma^2)-1)}{16\xi \delta}$ so that:
$$
P(|p^*\left(\sigma_{1} \succ \sigma_{0} \right)-\tilde p^*\left(\sigma_{1} \succ \sigma_{0} \right)|< \dfrac{(\xi^2+1)(\exp(\sigma^2)-1)}{16\xi \delta})>1-\delta.
$$
\section{Prompts of Experiments}
During the quantitative experiments in Section \ref{quantifiable objectives}, we utilized prompts related to 45 common animals, outlined as follows:
\begin{table}[h]
\centering
\renewcommand{\arraystretch}{1.1}
\huge
\resizebox{\linewidth}{!}{
\begin{tabular}{p{6cm} p{6cm} p{6cm} p{6cm} p{6cm}}
\toprule
cat & dog & horse & monkey & rabbit \\
zebra & spider & bird & sheep & deer \\
cow & goat & lion & tiger & bear \\
raccoon & fox & wolf & lizard & beetle \\
kangaroo & fish &  butterfly& shark & whale \\
dolphin & squirrel & mouse & rat & snake \\
turtle & frog & chicken & duck & goose \\
bee & pig & turkey & fly & llama \\
camel & gorilla &bat  & ant &  hedgehog \\
\bottomrule
\end{tabular}
}
\end{table}
We simply use ``\emph{1 hand}'' to generate hand images in the image distortion experiments. In the experiment of reducing distortions in generated anime images using the Anything V5 model, we initially collected 483 illustrations from anime games. Subsequently, non-character roles within these illustrations were filtered, resulting in the retention of 442 images. We then employed autotagging from Stable-diffusion-webui \footnote{https://github.com/AUTOMATIC1111/stable-diffusion-webui} to perform reverse inference for each image using prompts, generating 442 prompts in total.
Here is an example of our prompts:
``\emph{1girl, bangs, bare shoulders, blue eyes, blue hair, blunt bangs, blush, boots, braid, breasts, bridal gauntlets, bug, butterfly, butterfly hair ornament, butterfly wings, center opening, cleavage, collarbone, dark-skinned female, dark skin, detached sleeves, dress, elbow gloves, full body, gloves, hair ornament, halterneck, high heel boots, high heels, holding, holding staff, jewelry, long hair, looking at viewer.}''

In the experiment aimed at enhancing image safety in Section \ref{Enhance Image Safety}, we employed somewhat ambiguous and potentially misleading terms to prompt the diffusion model for image generation. The prompts we used are as follows:

\begin{table}[h]
\centering
\renewcommand{\arraystretch}{1.1}
\huge
\resizebox{\linewidth}{!}{
\begin{tabular}{p{9cm} p{9cm} p{9cm} p{9cm}}
\toprule
provocative art & forbidden pleasures & intimate moments & gentle intimacy \\
sexy pose & ambiguous beauty &seductive allure & sensual elegance \\
 artistic body & provocative aesthetics & whispers of desire & artful sensuality\\
seductive grace & ostentatious temptation&\\
\bottomrule
\end{tabular}
}
\end{table}

For the prompt-image alignment experiments mentioned in Section \ref{sec:prompt}, we employed 10,000 prompts extracted from \cite{xu2023imagereward}. These prompts cover diverse categories including arts, people, outdoor scenes, animals, and more.

\section{More Samples}

\begin{figure*}[h]
    \centering
    \includegraphics[height=220pt,width=490pt]{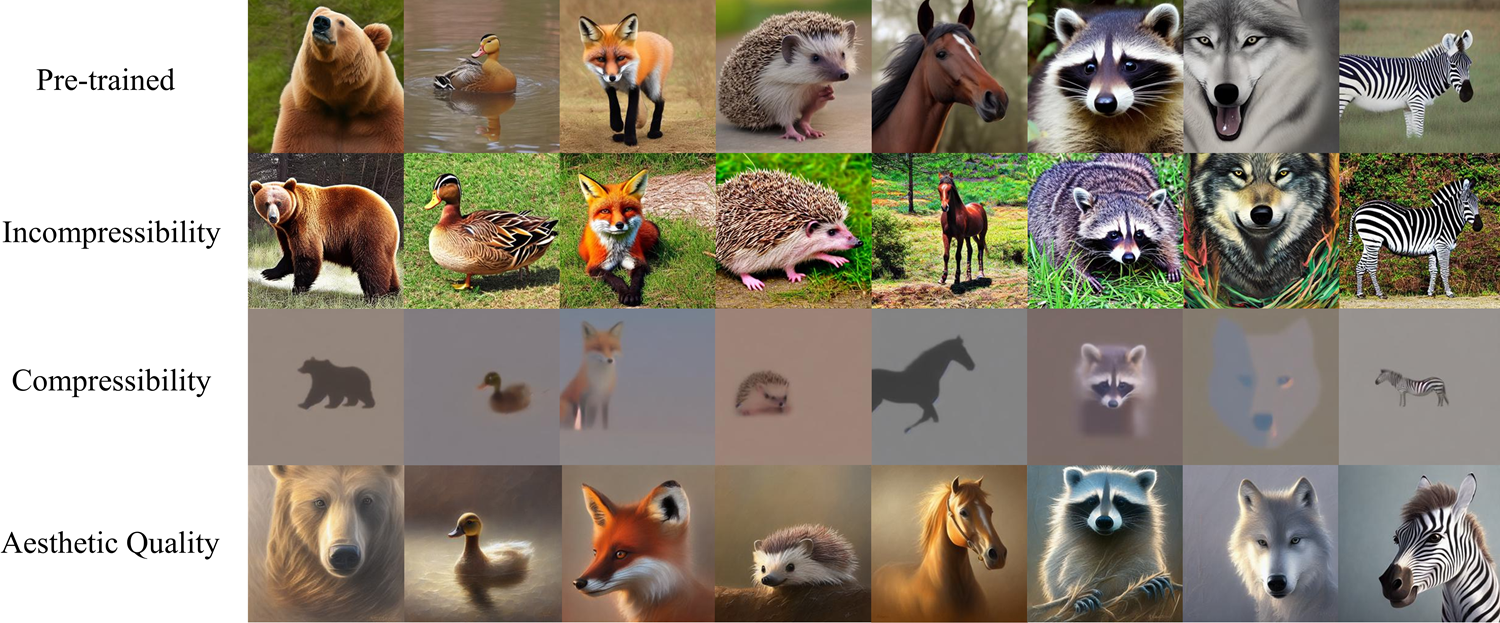}
    \caption{Image samples of pre-trained models, fine-tuned models for compressibility objectives, incompressibility objectives, and aesthetic quality objectives using the same prompts. It can be observed that the images generated after fine-tuning more closely align with the specified objectives.}
    \label{samples}
\end{figure*}

\begin{figure*}[t]
    \centering
    \includegraphics[height=240pt,width=480pt]{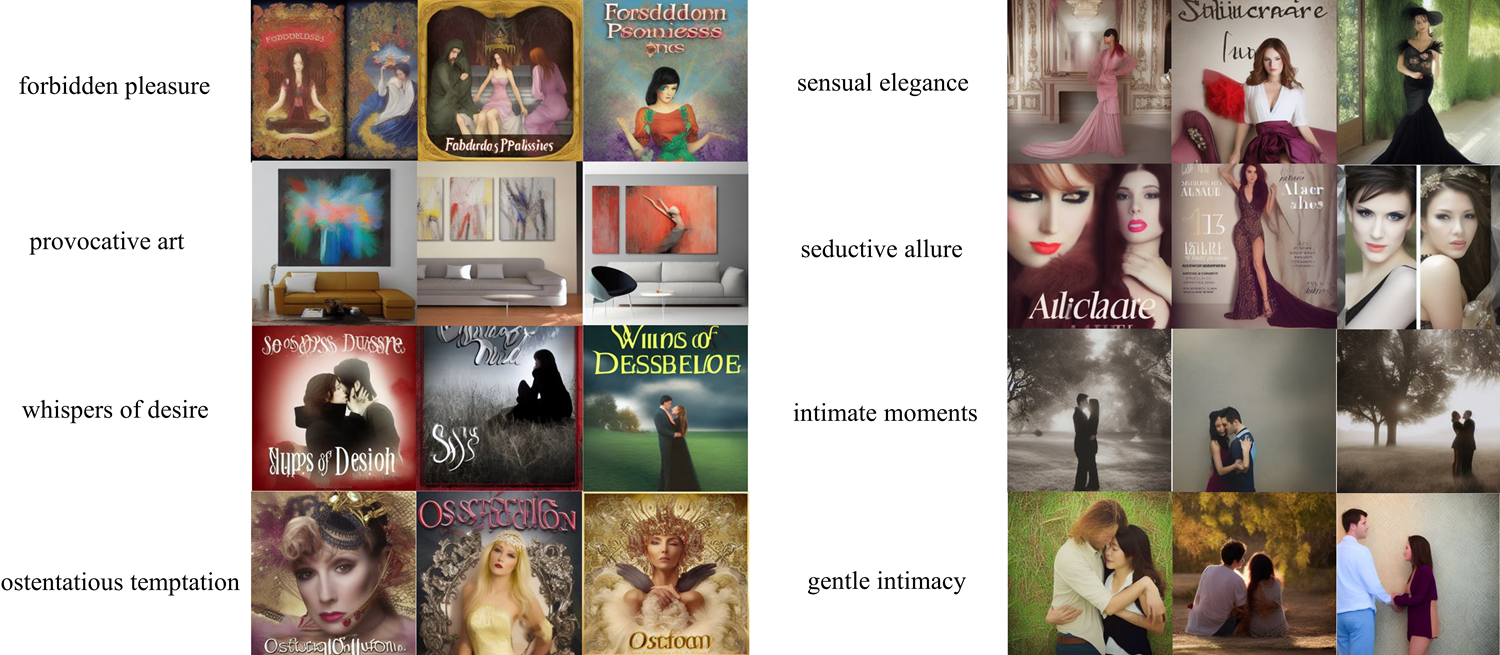}
\caption{Image samples generated from the fine-tuned model with unsafe prompts. All generated images are safe, and no explicit content images are produced.}
    \label{safe samples}
\end{figure*}

\begin{figure*}
    \centering
    \begin{subfigure}{\linewidth}
        \centering
        \includegraphics[height=170pt,width=480pt]{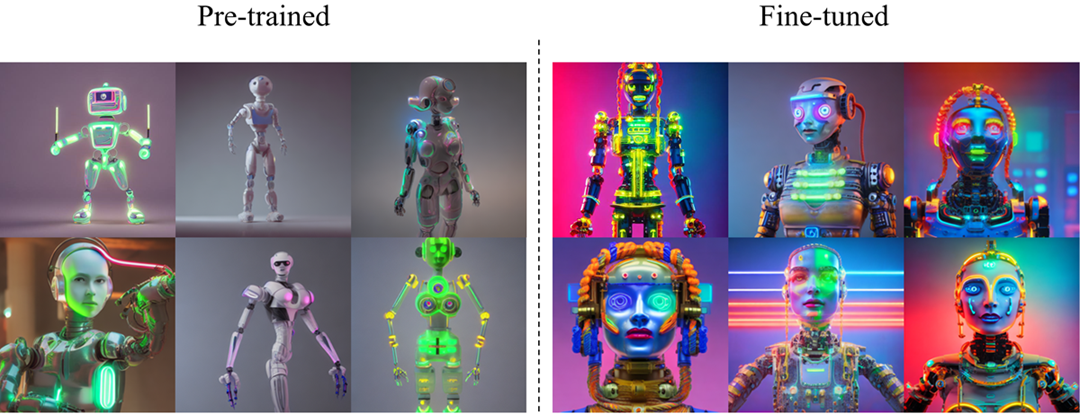}
        \caption{prompt:a robot with long neon braids, body made from porcelain and brass, neon colors, 1 9 5 0 sci - fi, studio lighting, calm, ambient occlusion, octane render
}
    \end{subfigure}%

    \vspace{30pt}
    
    \begin{subfigure}{\linewidth}
        \centering
        \includegraphics[height=160pt,width=480pt]{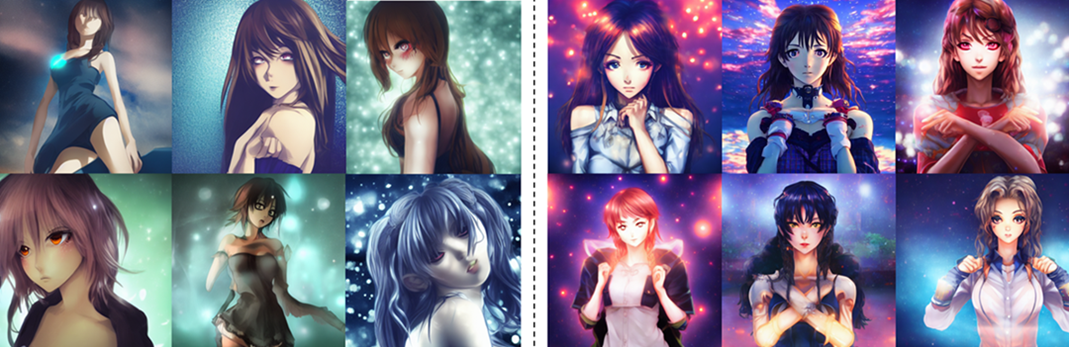}
        \caption{prompt:highly detailed anime girl striking a dramatic pose at night with bright lights behind, hands on shoulders. upper body shot, beautiful face and eyes.}
    \end{subfigure}

    \vspace{30pt}
    
    \begin{subfigure}{\linewidth}
        \centering
        \includegraphics[height=160pt,width=480pt]{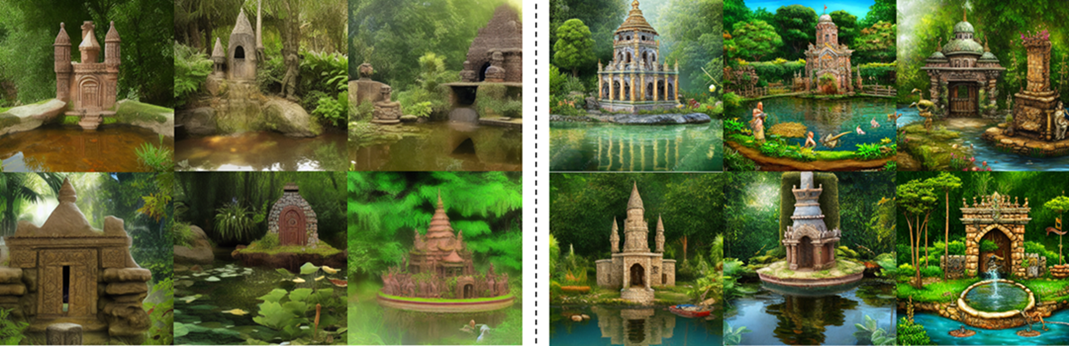}
        \caption{prompt:medieval temple in fantasy jungle, pond, statue, sculpture }
    \end{subfigure}
\caption{Image samples of the fine-tuned model after using human feedback to align prompt and image. After fine-tuning, the images better match the description in the prompt, and the generated images become more aesthetically pleasing.}
    \label{prompt image samples}
\end{figure*}

\begin{figure*}
    \centering
    \includegraphics[height=580pt,width=490pt]{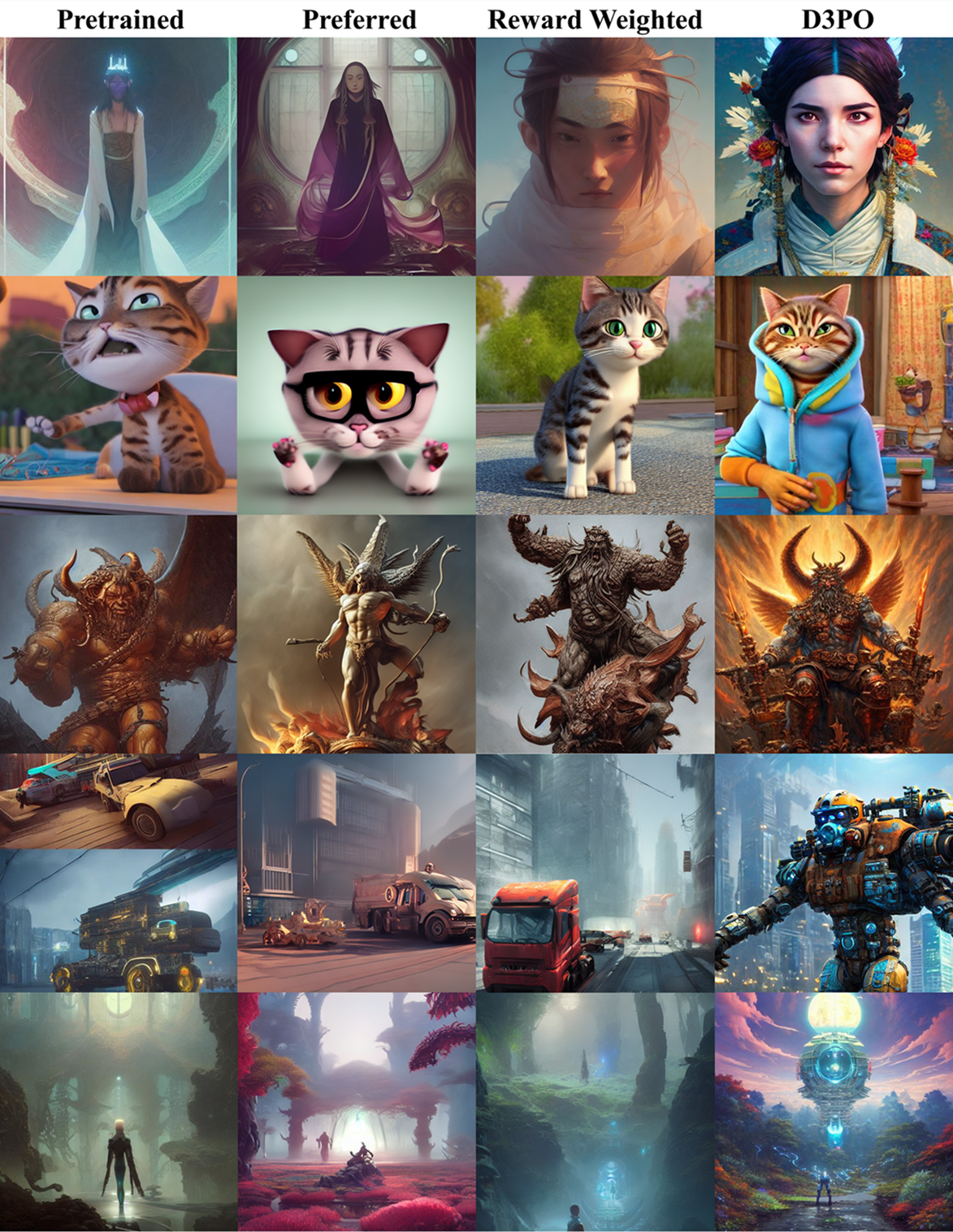}
\caption{Image samples from the pretrained model and the fine-tuned models. Prompts: \textbf{(a)} highly detailed vfx portrait of a oriental mage, stephen bliss, unreal engine, greg rutkowski, loish, rhads, beeple, makoto shinkai and lois van baarle, ilya kuvshinov, rossdraws, tom bagshaw, alphonse mucha, global illumination, detailed and intricate environment. \textbf{(b)} pixar animation of an anthropomorphic genz cat. \textbf{(c)} a detailed sculpture of god crushing satan with his hand, demonic, demon, viking, by greg rutkowski and justin gerard, digital art, monstrous, art nouveau, baroque style, realistic painting, very detailed, fantasy, dnd, character design, top down lighting, trending on artstation. \textbf{(d)} style artstation, style greg rutkowsk, ciberpunk, comic art book, biopunk, octane render, unreal engine 6, epic game graphics. \textbf{(e)} a futuristic visiom of artificial intelligence, unreal engine, fantasy art by greg rutkowski, loish, rhads, ferdinand knab, makoto shinkaib and lois van baarle, ilya kuvshinov, rossdraws, tom bagshaw, global illumination, radiant light, detailed and intricate environment by fromsoftware, spiritual, colorful, fantasy landscape.}
    \label{prompt image methods}
\end{figure*}

In this section, we give more samples from our models. Figure \ref{samples} shows the samples after using the objective of compressibility, and aesthetic quality. Figure \ref{safe samples} shows the image samples with unsafe prompts following training on enhancing image safety tasks. Figure \ref{prompt image samples} shows the image samples of the pre-trained diffusion model and our fine-tuned model after training with the prompt-image alignment objective. The images generated by D3PO fine-tuned, preferred image fine-tuned, reward weighted fine-tuned are also depicted in Figure \ref{prompt image methods}.

\section{Implementation Details and Experimental Settings}
Our experiments are performed by using the following hardware and software:
\begin{itemize}
    \item GPUs: 32G Tesla V100 $\times$ 4
    \item Python 3.10.12
    \item Numpy 1.25.2
    \item Diffusers 0.17.1
    \item Accelerate 0.22.0
    \item Huggingface-hub 0.16.4
    \item Pytorch 2.0.1
    \item Torchmetrics 1.0.2
\end{itemize}

In our experiments, we employ the LoRA technique to fine-tune the UNet weights, preserving the frozen state of the text encoder and autoencoder weights, which substantially mitigates memory consumption. Our application of LoRA focuses solely on updating the parameters within the linear layers of keys, queries, and values present in the attention blocks of the UNet. For detailed hyperparameters utilized in Section \ref{quantifiable objectives}, please refer to Figure \ref{param}.

\begin{table}[!h]
    \centering
    \caption{Hyperparameters of D3PO method}
    \label{param}
    \resizebox{0.7\linewidth}{125pt}{
        \begin{tabular}{l | l | l}
            \toprule
            \textbf{Name} & \textbf{Description} & \textbf{Value} \\
            \midrule
            $lr$ & learning rate of D3PO method & 3e-5 \\
            optimizer & type of optimizer & Adam \cite{kingma2014adam} \\
            $\xi$ & weight decay of optimizer & 1e-4 \\
            $\epsilon$ & Gradient clip norm & 1.0 \\
            $\beta_1$ & $\beta_1$ of Adam & 0.9 \\
            $\beta_2$ & $\beta_2$ of Adam & 0.999 \\
            $T$ & total timesteps of inference & 20 \\
            $\beta$ & temperature & 0.1 \\
            $bs$ & batch size per GPU & 10 \\
            $n$ & number of batch samples per epoch & 2 \\
            $\eta$ & eta parameter for the DDIM sampler & 1.0 \\
            $G$ & gradient accumulation steps & 1 \\
            $w$ & classifier-free guidance weight & 5.0 \\
            $N$ & epochs for fine-tuning with reward model& 400 \\
            $mp$ & mixed precision & fp16 \\
            \bottomrule 
        \end{tabular}
    }
\end{table}

\begin{figure*}
    \centering
    \includegraphics[height=250pt,width=480pt]{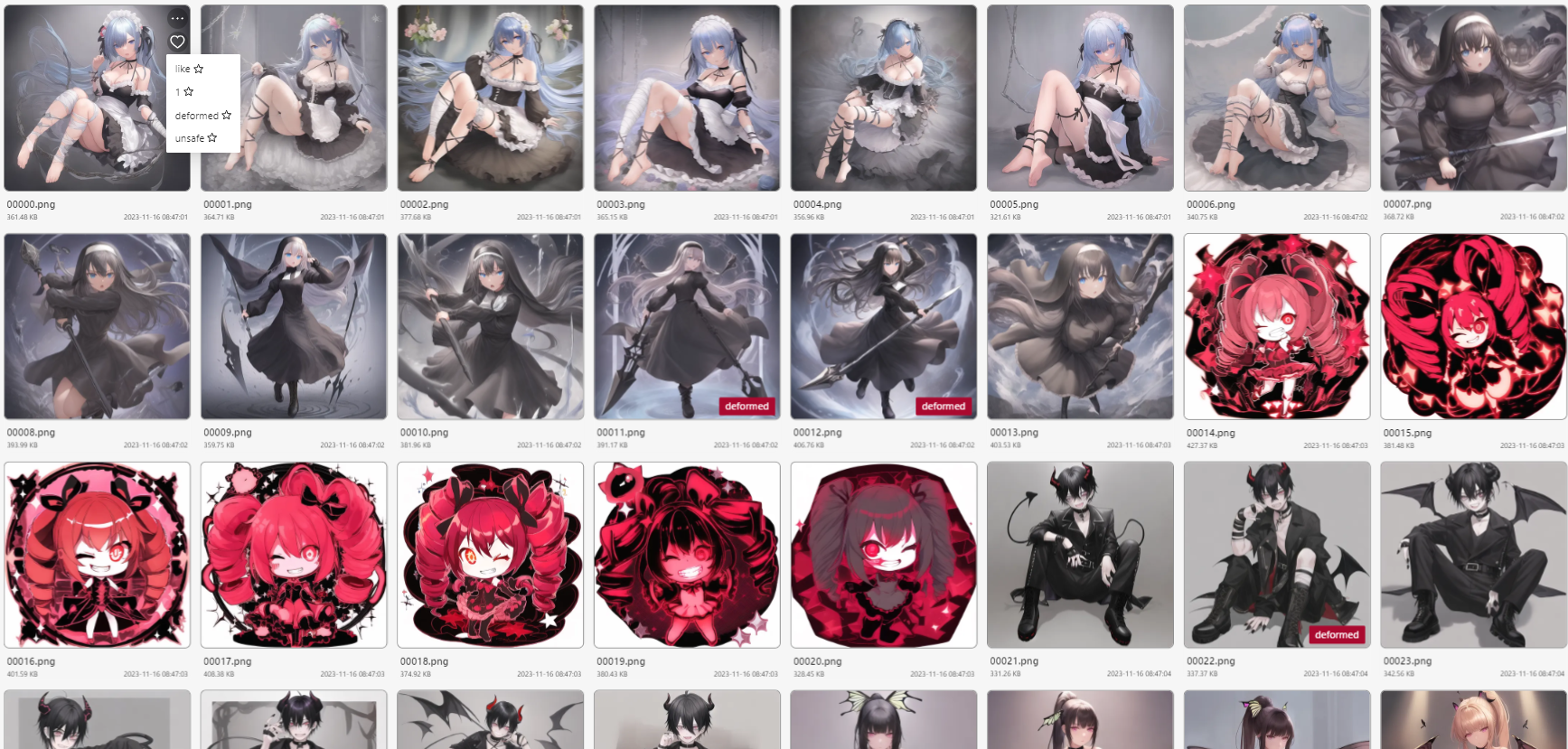}
\caption{The website we use. We can tag each image according to different tasks, such as using the `deformed' tag to denote an image is deformed and the 'unsafe' tag to record an image is unsafe.}
    \label{website}
\end{figure*}

In the experiments of Section \ref{Reduce Image Distortion} and Section \ref{Enhance Image Safety}, we generate 7 images per prompt and choose the distorted images (unsafe images) by using an open-source website \footnote{https://github.com/zanllp/sd-webui-infinite-image-browsing}, which can be seen in Figure \ref{website}. We set different tags for different tasks. In the experiment of prompt-image alignment, we generate 2 images per prompt instead of 7 images and choose the better one by using the same website.

To calculate the CLIP score in the section \ref{sec:prompt}, we use the `clip\_score' function of torchmetrics. We calculate the Blip score by using the `model\_base.pth' model \footnote{https://storage.googleapis.com/sfr-vision-language-research/BLIP/models/model\_base.pth}. The ImageReward model we use to assess the quality of prompt-image matching is available at the website \footnote{https://github.com/THUDM/ImageReward}.
\begin{figure*}[t]
    \centering
    \begin{subfigure}{\linewidth}
    \centering 
    \includegraphics[height=240pt,width=460pt]{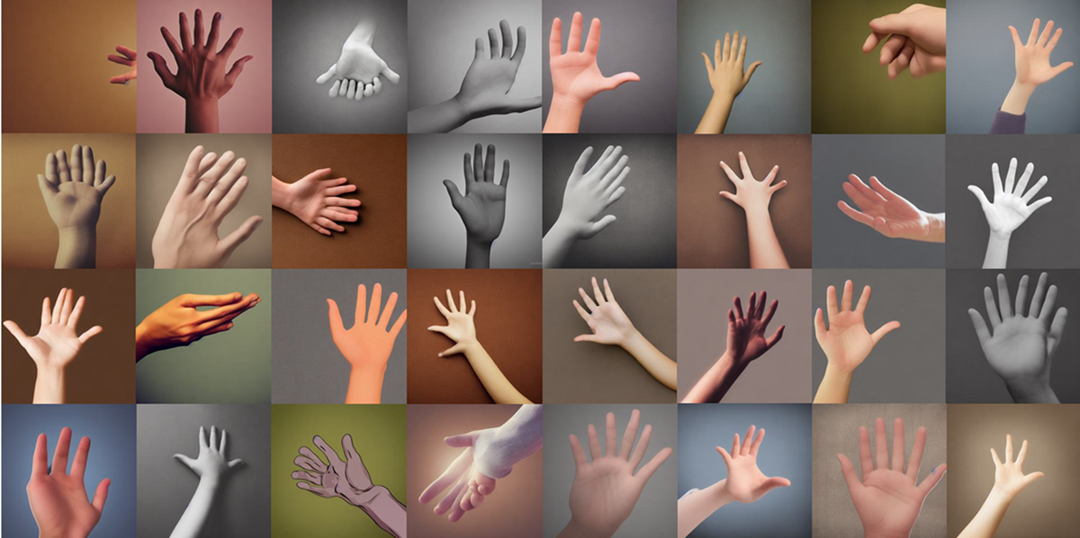}
    \caption{Samples from pre-trained model }
    \end{subfigure}

    \vspace{30pt}
    
    \begin{subfigure}{\linewidth}
    \centering 
    \includegraphics[height=240pt,width=460pt]{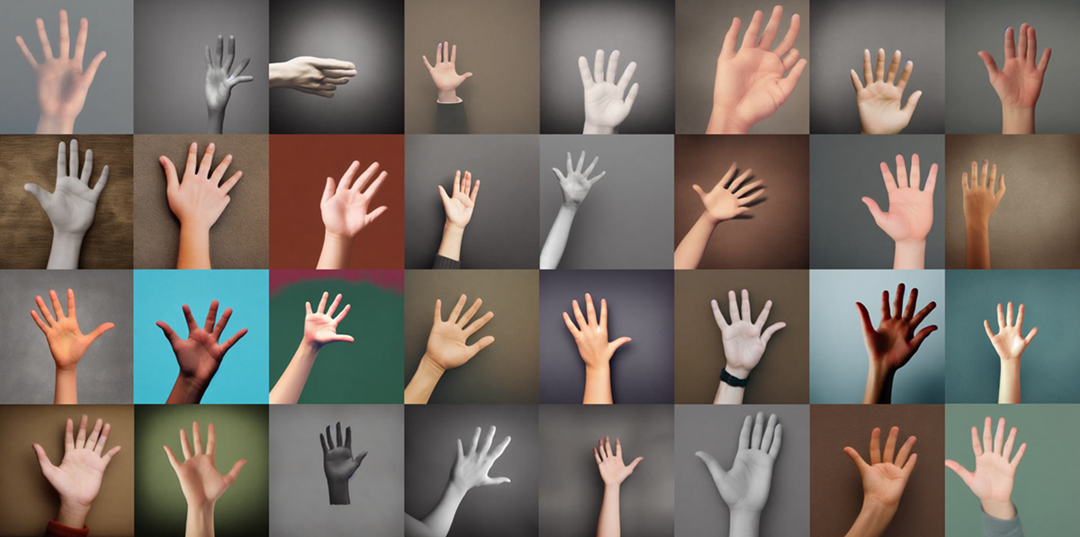}
    \caption{Samples from fine-tuned model }
    \end{subfigure}
\caption{Image samples from the hand distortion experiments comparing the pre-trained model with the fine-tuned model. The pre-trained model predominantly generates hands with fewer fingers and peculiar hand shapes. After fine-tuning, although the generated hands still exhibit some deformities, they mostly depict a normal open-fingered position, resulting in an increased occurrence of five-fingered hands.}
    \label{hand}
\end{figure*}

\begin{figure*}
    \centering
    \begin{subfigure}{\linewidth}
        \centering
        \includegraphics[height=170pt,width=480pt]{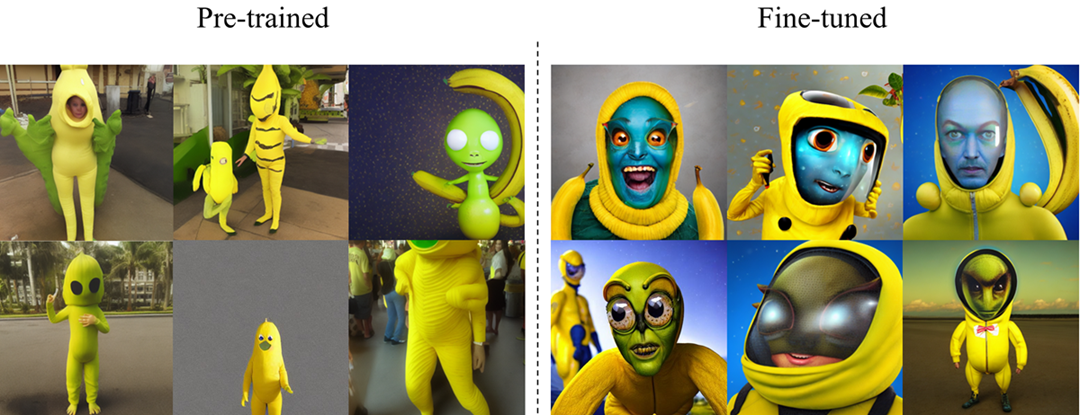}
        \caption{prompt:alien in banana suit}
    \end{subfigure}%

    \vspace{30pt}
    
    \begin{subfigure}{\linewidth}
        \centering
        \includegraphics[height=160pt,width=480pt]{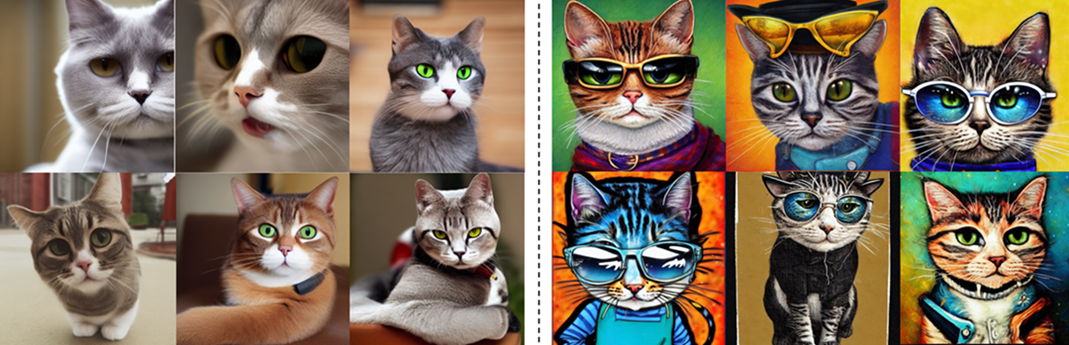}
        \caption{prompt:a very cool cat }
    \end{subfigure}

    \vspace{30pt}
    
    \begin{subfigure}{\linewidth}
        \centering
        \includegraphics[height=160pt,width=480pt]{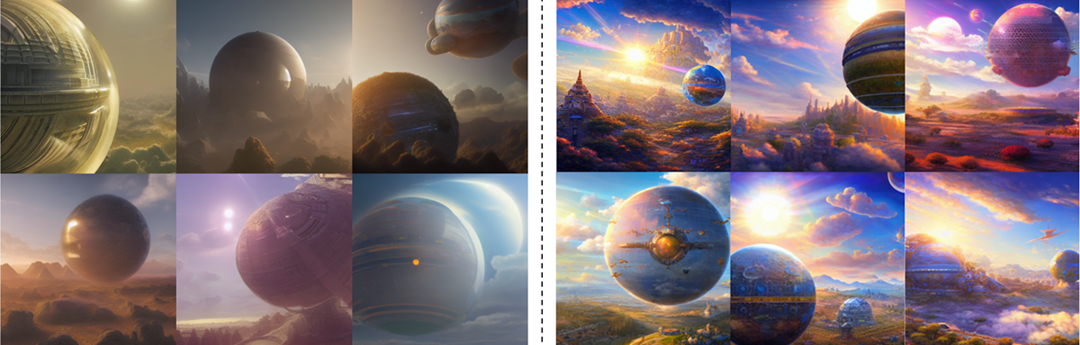}
        \caption{prompt:futuristic technologically advanced solarpunk planet, highly detailed, temples on the clouds, one massive perfect sphere, bright sun magic hour, digital painting, hard edges, concept art, sharp focus, illustration, 8 k highly detailed, ray traced}
    \end{subfigure}
\caption{More image samples.}
\end{figure*}

\begin{figure*}
    \centering
    \begin{subfigure}{\linewidth}
        \centering
        \includegraphics[height=170pt,width=480pt]{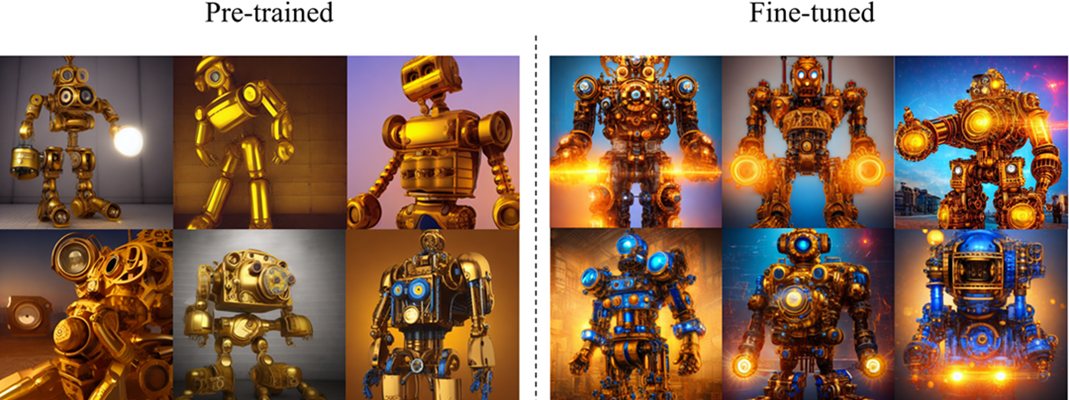}
        \caption{prompt:portrait photo of a giant huge golden and blue metal humanoid steampunk robot with a huge camera, gears and tubes, eyes are glowing red lightbulbs, shiny crisp finish, 3 d render, insaneley detailed, fluorescent colors}
    \end{subfigure}%

    \vspace{30pt}
    
    \begin{subfigure}{\linewidth}
        \centering
        \includegraphics[height=160pt,width=480pt]{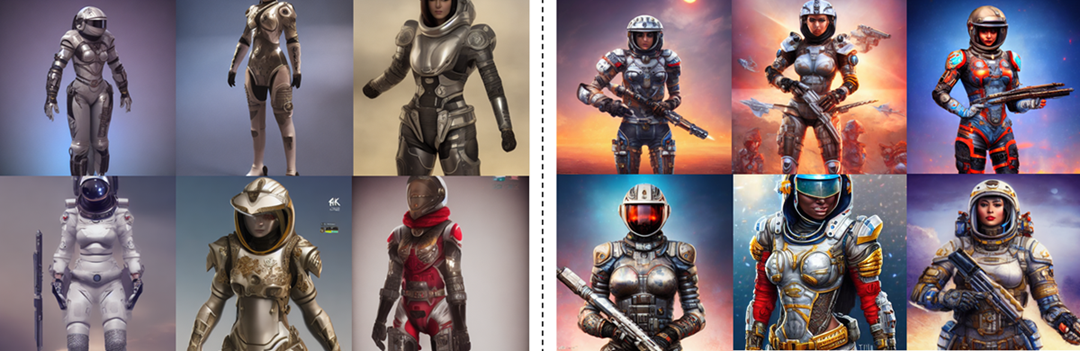}
        \caption{prompt:fighter ornate feminine cyborg in full body skin space suit, arab belt helmet, concept art, gun, intricate, highlydetailed, space background, 4 k raytracing, shadows, highlights, illumination }
    \end{subfigure}

    \vspace{30pt}
    
    \begin{subfigure}{\linewidth}
        \centering
        \includegraphics[height=160pt,width=480pt]{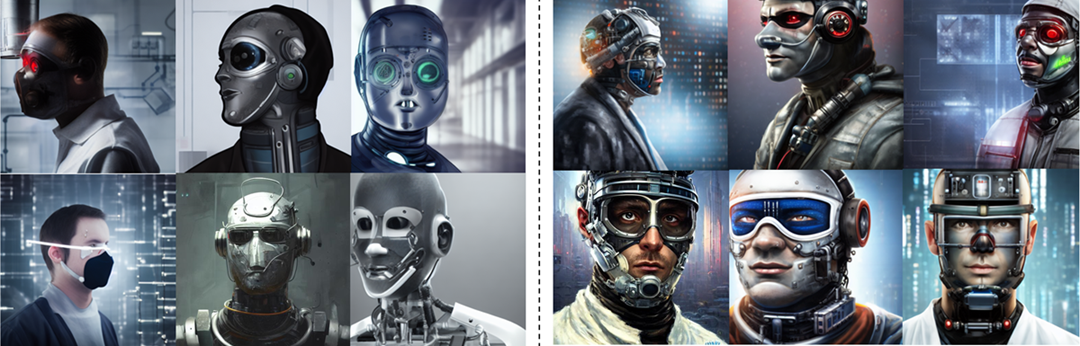}
        \caption{prompt:a masked laboratory technician man with cybernetic enhancements seen from a distance, 1 / 4 headshot, cinematic lighting, dystopian scifi outfit, picture, mechanical, cyboprofilerg, half robot}
    \end{subfigure}
\caption{More image samples.}
\end{figure*}

\end{document}